\documentclass{article}
\PassOptionsToPackage{numbers,sort&compress}{natbib}
\usepackage[eandd, final]{neurips_2026}

\usepackage[utf8]{inputenc} 
\usepackage[T1]{fontenc}    
\usepackage{hyperref}       
\usepackage{etoc}
\usepackage{url}            
\usepackage{booktabs}       
\usepackage{amsfonts}       
\usepackage{nicefrac}       
\usepackage{microtype}      
\usepackage[table]{xcolor}         
\usepackage{placeins}
\usepackage[normalem]{ulem}

\usepackage{amssymb} 
\newcommand{\cmark}{\checkmark}
\newcommand{\xmark}{\texttimes}

\usepackage{multirow}
\usepackage{graphicx}
\usepackage{tabularx}
\usepackage{caption}
\usepackage{colortbl, array}
\usepackage{wrapfig,lipsum,booktabs}
\usepackage{subcaption}
\usepackage{rotating}
\usepackage{adjustbox}
\usepackage{pgfplots}
\pgfplotsset{compat=1.18}
\usepackage{tikz}
\usepackage{float}
\usepackage{wrapfig,lipsum,booktabs}
\definecolor{improved}{RGB}{255,235,200}
\usepackage{amsmath}
\usepackage{cleveref}

\usepackage[most]{tcolorbox}
\newtcolorbox{promptbox}[1]{
  enhanced, breakable,
  colback=gray!5, colframe=gray!60!black,
  fonttitle=\bfseries, title=#1,
  boxrule=0.6pt, arc=2pt,
  left=6pt, right=6pt, top=4pt, bottom=4pt
}

\newcommand{\SA}[1]{{\color{magenta}SA: #1}}

\definecolor{darkgreen}{RGB}{0,100,0}

\newcommand{\DS}[1]{\color{red}DS: #1}
\newcommand{\SV}[1]{{\color{violet} {\bf} SV: #1}}

\definecolor{overall}{RGB}{220,235,255} 
\definecolor{best}{RGB}{180,215,180}      
\definecolor{second}{RGB}{255,245,200}    

\title{MaRiO: Multi-agent Collaborative Reasoning via Shared Observations in MLLMs}

\author{
  Nathaniel Redmond\thanks{Equal contribution.} \quad
  Fidel Omar Tito Cruz\footnotemark[1] \quad
  Devansh Sharma \\
  \textbf{Shehreen Azad \quad Sirshapan Mitra \quad Shruti Vyas \quad Yogesh S. Rawat} \\[4pt]
  Institute of Artificial Intelligence, University of Central Florida \\
}

\begin{document}

\maketitle
\etocdepthtag.toc{mtchapter} 

\begin{abstract}
Multimodal Large Language Models (MLLMs) show strong visual perception capabilities, but existing evaluations largely focus on single-view or multi-view settings with shared camera parameters, leaving the challenge of {multi-agent collaborative} scenarios underexplored. In such settings, multiple agents observe a shared environment from independent viewpoints, and reliable decisions require reasoning across these independent observations of the same environment. To study this problem, we introduce \textbf{MaRiO}, a benchmark for evaluating multi-agent collaborative reasoning, and evaluate \textbf{23} state-of-the-art MLLMs on it. Our analysis reveals three key findings: (1) tasks governed by static perception are generally easier than tasks that require transformational event sequences; (2) most models struggle to resolve cross-agent correspondences, with errors increasing with scene complexity; and (3) models show limited ability to leverage geometric cues present in the scene. As an initial step toward addressing these challenges, we explore geometry-guided synthetic view augmentation that generates intermediate transition frames between agent observations, providing auxiliary spatial context that improves performance in certain categories. Overall, our results highlight that multi-agent collaborative reasoning remains an open challenge and an important direction for multimodal embodied systems. The code is available on the project page: \href{https://omarftt.github.io/MaRiO-page/}{https://omarftt.github.io/MaRiO-page/}.
\end{abstract}

\section{Introduction}
\label{sec:intro}
\begin{figure}[t!]
    \centering
    \includegraphics[width=.96\linewidth]{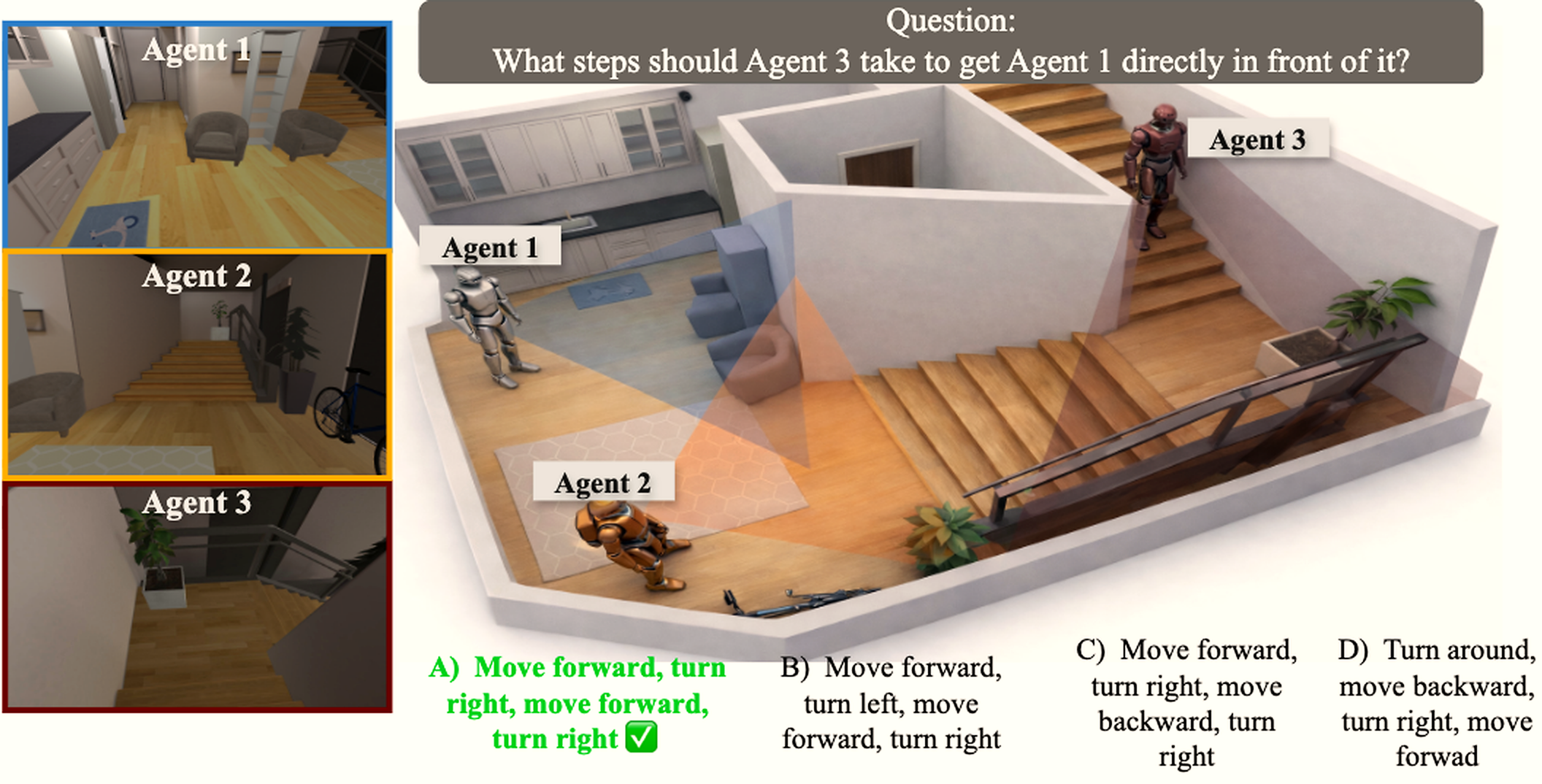}
    \caption{\textbf{Multi-agent collaborative reasoning in a shared environment,} where multiple agents observe the same environment from independent coordinate frames, requiring cross-agent correspondence before inferring spatial and geometric relations.
    }
    \label{fig:overview}
\end{figure}

Multimodal Large Language Models (MLLMs) achieve strong performance across diverse visual tasks \cite{zhu2025internvl3, wang2024qwen2vl, zhang2024llavavideo, zhang2024llavanextvideo, azad2025hierarq}. To assess whether this reflects genuine spatial comprehension, recent works evaluate spatial and geometric perception abilities like layout, position, orientation, viewpoint and depth relations \cite{schiappa2024probing, tong2024cambrian,zhang2025spinbench,zhang2025do,thrush2022winoground,azad2025understanding}. These capabilities are essential for embodied applications, like robotic manipulation \cite{Chen_2024_CVPR,song2025robospatial,feng2025seeingviewsbenchmarkingspatial}, autonomous navigation \cite{anderson2018vision,shridhar2020alfred}, and collaborative augmented reality \cite{grauman2022ego4d}, where agents must reason about space to plan and act safely. In collaborative environments, multiple agents observe the same scene from different viewpoints \cite{zhang2023building, zhang2024combo} and must coordinate their actions (\Cref{fig:overview}).

Prior work on spatial understanding primarily focus on single-view settings \cite{du2024embspatial,cheng2024spatialrgpt,tong2024cambrian,jia2026omnispatial,yang2025thinking} limited to one perspective and cannot resolve cross-view ambiguities. Multi-view approaches \cite{feng2025seeingviewsbenchmarkingspatial,yeh2025seeing,gholami2025spatialreasoningvisionlanguagemodels,yin2025spatialmentalmodelinglimited,yang2025mmsi,li2025viewspatialbenchevaluatingmultiperspectivespatial} extend this by evaluating spatial consistency across multiple images, typically captured along a shared trajectory with known or recoverable geometric relationships. 

Therefore, they do not require joint reasoning over distances, motion, or navigational constraints. In contrast, multi-agent settings \cite{gholami2025spatialreasoningvisionlanguagemodels} introduce an additional challenge: each agent observes the environment from its own egocentric frame without shared geometric metadata. Such settings can arise in multi-robot teams without known relative camera poses, drone teams operating in GPS-denied environments, or collaborative AR/VR systems without a common spatial anchor. As a result, models must first establish cross-agent correspondences between independent viewpoints and align them before spatial and geometric relationships can be inferred, a capability necessary for navigation, collision avoidance, and coordinated movement in collaborative environments.

To study multi-agent collaborative reasoning capabilities of MLLMs, we introduce \textbf{MaRiO}, a benchmark based on \textbf{M}ultiple \textbf{a}gents \textbf{R}easoning from \textbf{I}ndependent \textbf{O}bservations of a shared environment. Unlike prior single-view or multi-view setups, MaRiO simulates scenarios where multiple agents independently observe the same environment from different viewpoints without access to others' camera poses, coordinate frames, or scene geometry. The benchmark is built around 5 tasks: Distance Awareness, Occlusion, Relative Positioning, Egocentric Motion, and Navigation, collectively requiring both static spatial perception and compositional reasoning over geometric transformations. By requiring models to infer relationships purely from independently observed views, MaRiO introduces a two-stage challenge: establishing cross-agent correspondences and reasoning over the aligned spatial relationships.

We evaluate \textbf{23} recent open-source and proprietary MLLMs on MaRiO and find that all models perform substantially below human level, highlighting the difficulty of multi-agent collaborative reasoning. Our analysis shows that many failures arise because models struggle to reliably establish cross-agent correspondences between independent viewpoints. Importantly, this challenge stems not simply from the number of observations but from the need to infer relationships between viewpoints observed from different perspectives. To mitigate this issue, we explore  geometry-guided synthetic intermediate observations that bridge available agent views. These transition views are generated from reconstructed geometry and incorporated as auxiliary visual context for the MLLM. This approach helps reveal spatial transitions between viewpoints and improves performance in several multi-agent collaborative reasoning tasks.

Our contributions are as follows:

\begin{itemize}
    \item We introduce \textbf{MaRiO}, a multi-agent benchmark for studying multi-agent collaborative reasoning across independently observed views that requires models to establish cross-agent correspondence before reasoning.
    \item We conduct a large-scale evaluation of \textbf{23} state-of-the-art open-source and proprietary MLLMs, revealing substantial performance gaps compared to human reasoning and providing detailed insights into their limitations.
    \item We further explore diffusion-guided synthetic view augmentation to generate transition views between agent observations for improving multi-agent collaborative reasoning.
\end{itemize}

\section{Related Works}
\label{sec:rel_work}
\label{sec:rel_work}

\noindent \textbf{Spatial Reasoning Benchmarks.}  Recent MLLMs have shown expanded capabilities across visual reasoning tasks~\cite{bai2025qwen25vltechnicalreport,li2024llava,lu2024ovis,abdin2024phi3technicalreporthighly,zhu2025internvl3,team2023gemini,liu2025oryxmllmondemandspatialtemporal}. Recent efforts further improve spatial understanding through temporally enriched inputs, spatial encoders, and 3D-aware representations \cite{wu2025spatial,fan2025vlm,wang2025vggt,wang2025continuous,li2025viewspatialbenchevaluatingmultiperspectivespatial,vst}. However, it remains unclear whether these models can reason across independently oriented agents observing the same environment. Existing spatial benchmarks primarily evaluate single-view compositional, relational, 3D grounding, and embodied reasoning tasks ~\cite{hudson2019gqa,suhr2019corpus,yi2019clevrer,liu2024mmbench,fu2023mme,li2023seed,fu2024blink,li-etal-2023-evaluating,thrush2022winoground,liu2023visual,ma20243dsrbench,jia2026omnispatial,das2018embodied,shridhar2020alfred,anderson2018vision,chen2020scanrefer,azuma2022scanqa,ma2022sqa3d}, often with geometric metadata such as scene graphs, camera intrinsics, or 3D annotations. Multi-view and egocentric benchmarks ~\cite{grauman2022ego4d,gholami2025spatialreasoningvisionlanguagemodels,yeh2025seeing,li2025viewspatialbenchevaluatingmultiperspectivespatial,zhang2025spinbench,xue2025reasoning, yang2025mmsi} extend evaluation to cross-view consistency, but typically assume shared trajectories, recoverable geometry, or observations drawn from a common 3D context. Meanwhile, existing multi-agent benchmarks ~\cite{xu2025vs,duan2024gtbench,huang2025competing} mainly study strategic decision-making in language-only settings. In contrast, MaRiO evaluates multimodal multi-agent collaborative reasoning across independently observed views, where models must establish cross-agent correspondence without shared geometric metadata.

\noindent \textbf{View Synthesis and Geometry-Guided Generation.}
Prior work recovers missing viewpoints using 3D reconstruction methods such as point clouds \cite{wang2024dust3r, leroy2024grounding, wang2025vggt} and Gaussian splatting \cite{kerbl20233d, huang20242d}; but most MLLMs cannot directly process these modalities without costly architectural changes. Inspired  by diffusion-based novel view synthesis from reconstructed geometry \cite{yu2024viewcraftertamingvideodiffusion, liu2024sora, wang2024motionctrlunifiedflexiblemotion}, we generate intermediate viewpoints that reveal spatial transitions between agent observations that can aid multi-agent collaborative reasoning without modifying the underlying MLLM.
\section{Benchmarking Multi-agent Collaborative Reasoning}
\label{sec:benchmark}

\begin{table}[t!]
    \centering
    \caption{\textbf{Comparison of MaRiO with existing visual reasoning benchmarks.} MV and MA respectively denote Multi-view and Multi-agent. Perc.\ and Comp.\ respectively refer to static perception and compositional reasoning over geometric transformations. Cam. Geom. refers to the provision of explicit camera geometry, and Aux. Info. refers to auxiliary information beyond RGB frames.}
    \setlength{\tabcolsep}{3pt}
    \resizebox{\linewidth}{!}{
    \begin{tabular}{l|cc|cccc|c}
    \toprule
        \multirow{2}{*}{\textbf{Benchmark}} & \multicolumn{2}{c|}{\textbf{Setup}} & \multicolumn{4}{c|}{\textbf{Reasoning Requirement}} & \multirow{2}{*}{\textbf{Reasoning Challenge}}\\
        & \textbf{MV} &\textbf{MA} & \textbf{Perc.} & \textbf{Comp.} & \textbf{Cam. Geom.} & {\textbf{Aux. Info.}} \\
        \midrule
        VSR \cite{liu2023visual} & \xmark & \xmark & \cmark & \xmark&-& None & Spatial relation \\
        Winoground \cite{thrush2022winoground} & \xmark & \xmark & \cmark & \xmark&-& Attribute tag & Compositionality\\
        GeoMeter \cite{azad2025understanding} &  \xmark & \xmark & \xmark & \cmark&-& None & Depth perception\\
        OmniSpatial \cite{jia2026omnispatial} & \xmark & \xmark & \cmark & \cmark &-& None & Cognitive spatial reasoning\\
        ScanQA\cite{azuma2022scanqa} & \xmark & \xmark & \cmark & \xmark &- & Point Clouds & 3D scene understanding\\
        3DSRBench \cite{ma20243dsrbench} & \cmark & \xmark & \cmark & \cmark & Provided & None & 3D spatial reasoning \\
        ViewSpatial \cite{li2025viewspatialbenchevaluatingmultiperspectivespatial} & \cmark & \xmark & \cmark & \xmark & Provided & 3D Coord. & 3D scene understanding\\
        SpinBench \cite{zhang2025spinbench} & \cmark & \xmark & \cmark & \xmark & Provided & Cam. Intrinsics & Spatial transformation\\
        All-Angles\cite{yeh2025seeing} & \cmark & \xmark & \cmark & \cmark & Provided & Cam. Intrinsics & Cross-view correspondence\\
        VSI-Bench \cite{yang2025thinking} & \cmark & \xmark & \cmark & \cmark & Provided & None & Visual-spatial intelligence \\
        MMSI-Bench \cite{yang2025mmsi} & \cmark & \xmark & \cmark & \cmark & Provided & None & Multi-image spatial reasoning \\
        \midrule
        \textbf{MaRiO (Ours)} & \textbf{\cmark} & \textbf{\cmark}  & \textbf{\cmark} &  \textbf{\cmark} & \textbf{Not Provided} & \textbf{None} & \textbf{Cross-agent correspondence}\\
        
         \bottomrule
    \end{tabular}
    }
    \label{tab:bench_comp}
\end{table}

\begin{figure}[t!]
    \centering
    \includegraphics[width=\textwidth]{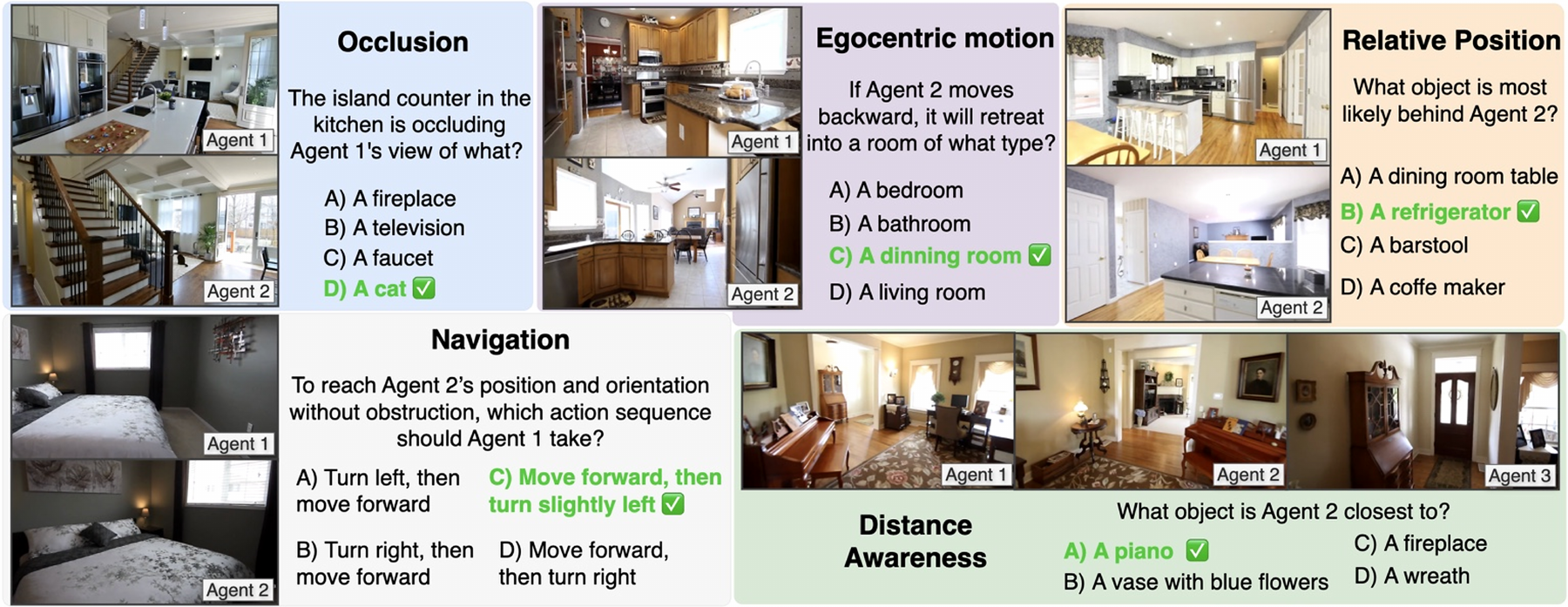}
    \caption{\textbf{Overview of MaRiO.} The benchmark consists of five spatial reasoning tasks: Occlusion, Egocentric Motion, Relative Position, Navigation, and Distance Awareness; evaluated in a multiple-choice question (MCQ) format. Each question is constructed using observations from multiple agents viewing the same environment, requiring reasoning across the images to determine the correct answer, without access to auxiliary information such as camera parameters or geometric metadata.}
    \label{fig:task_example}
\vspace{-15pt}    
\end{figure}

We introduce MaRiO, a multiple-choice VQA benchmark for collaborative multi-agent reasoning over independently oriented viewpoints without shared geometric metadata (Table~\ref{tab:bench_comp}). It contains 3{,}255 QA samples across 1{,}680 unique observation sets, each comprising views from multiple agents within a shared environment (Figure~\ref{fig:task_example}).

\subsection{MaRiO Construction}
MaRiO contains 1,680 unique collaborative multi-agent observation sets, distributed as 1,371 pairs and 309 triplets of views, forming a total of 3,425 unique observations. Each set captures synchronized views of a shared scene from independently oriented agents. 

\noindent \textbf{Task Definition.} MaRiO targets five reasoning dimensions across two categories: {Static Perception} focuses on spatial relationships and visibility, covering (1)~\textit{Occlusion}, which tests viewpoint-dependent visibility and object enumeration under partial occlusion; 
(2)~\textit{Relative Position}, which requires inferring relative agent positions and orientations; and (3)~\textit{Distance Awareness}, which tests qualitative depth and proximity estimation from an egocentric viewpoint.
Compositional Reasoning focuses on sequencing geometric transformations and includes (4)~\textit{Egocentric Motion}, which tests orientation under reference-frame transformations and (5)~\textit{Navigation}, which requires sequential state updates through compositions of translational and rotational actions.

\noindent \textbf{Image Generation.} MaRiO is constructed using both real-world images and synthetic simulators to balance realism and controllability. For real-world images, we incorporate 200 scenes from RealEstate10K \cite{zhou2018stereo}, which provides unconstrained house-tour videos with natural viewpoint variation, lighting, and object layouts, and 35 scenes from NAVI \cite{jampani2023navi}, offering controlled multi-perspective object observations. These sources contribute a total of 913 real-world image sets. To ensure coverage of targeted spatial phenomena, we further generate 39 scenes using VirtualHome \cite{puig2018virtualhomesimulatinghouseholdactivities}, Habitat-Sim \cite{puig2023habitat3}, and AI2-THOR \cite{ai2thor} simulators, yielding 767 synthetic image sets.

Some synthetic images are augmented with lighting variations, minor object perturbations, or lens artifacts to increase visual diversity.

\noindent\textbf{QA Generation.} QA pairs were manually generated through a structured annotation process involving three domain-expert annotators. Initial questions were jointly drafted, reviewed for clarity and spatial validity, and cross-verified by an additional annotator. Each reasoning category followed predefined specifications defining the spatial primitives, reference-frame assumptions, and reasoning constraints to ensure consistency. Questions were designed so that integrating information across views was necessary to determine the correct answer. All instances follow a four-choice multiple-choice format with plausible distractors and randomized answer-option order to avoid any inherent position bias. Exact prompts and additional examples are provided in the Supplementary. An LLM was used as a writing assistant during this process — specifically, to polish question phrasing and to draft initial candidate distractors; all outputs were reviewed, edited, and accepted or rejected by the human annotators, and ground-truth labels were authored by humans only.

\begin{table}[t!]
\centering
\caption{\textbf{Zero-shot performance of multi-agent collaborative reasoning on MaRiO.} Dist.: Distance awareness, Ego.: Egocentric motion, Nav.: Navigation, Occ.: Occlusion, and Rel.: Relative position. Model type-wise \colorbox{best}{best} and \colorbox{second}{second-best} scores are highlighted, with  overall \textbf{best} and \underline{second-best} additionally emphasized. Comp. and Static Perc. respectively refer to Compositional Reasoning over Geometric Transformations and Static Perception.}
\label{tab:main_results}

\setlength{\tabcolsep}{3pt}
\renewcommand{\arraystretch}{0.92}
{\small
\begin{tabular}{l|cc|ccc|c|c}
    \toprule
    \multirow{2}{*}{\textbf{Model}} & \multicolumn{2}{c|}{\textbf{Comp.}} & \multicolumn{3}{c|}{\textbf{Static Perc.}} & \multirow{2}{*}{\textbf{Overall}} & \multirow{2}{*}{\textbf{Avg.}}\\
    & \textbf{Ego.} & \textbf{Nav.} & \textbf{Dist.} & \textbf{Occ.} & \textbf{Rel.} & & \\
\midrule
Human Evaluators & 64.7 & 68.8 & 86.7 & 77.8 & 68.8 & 73.2 & 73.3 \\
Random Choice & 21.7 & 24.9 & 22.1 & 26.5 & 25.3 & 23.9 & 24.1 \\
\midrule
\rowcolor{blue!5}\multicolumn{8}{l}{\textbf{Proprietary Models}} \\
GPT 4o~\cite{achiam2023gpt} & 26.4 & 41.8 & 43.7 & 51.1 & 34.1 & 38.3 & 39.4 \\
GPT 5.1~\cite{singh2025openai} & \cellcolor{second}\underline{32.3} & 43.9 & \cellcolor{second}{47.4} & \cellcolor{second}\underline{53.6} & \cellcolor{second}\underline{55.9} & \cellcolor{second}\underline{45.2} & \cellcolor{second}\underline{46.6} \\
Gemini-2.5-Flash~\cite{team2023gemini} & 31.2 & \cellcolor{second}\underline{44.7} & 46.9 & 49.4 & 46.3 & 42.8 & 43.7 \\
Gemini-2.5-Pro~\cite{team2023gemini} & \cellcolor{best}\textbf{49.5} & \cellcolor{best}\textbf{52.1} & \cellcolor{best}\textbf{58.4} & \cellcolor{best}\textbf{60.3} & \cellcolor{best}\textbf{63.4} & \cellcolor{best}\textbf{55.7} & \cellcolor{best}\textbf{56.7} \\
\midrule
\rowcolor{blue!5}\multicolumn{8}{l}{\textbf{Open-Source Spatial Models}} \\
Spatial-MLLM-4B~\cite{wu2025spatial} & \cellcolor{second}{28.8} & \cellcolor{second}{32.5} & \cellcolor{best}{45.2} & \cellcolor{second}{40.6} & 37.4 & \cellcolor{second}{35.9} & \cellcolor{second}{36.9} \\
VST-7B-RL~\cite{vst} & 28.3 & 32.8 & 41.6 & 38.9 & \cellcolor{second}{38.3} & 35.1 & 36.0 \\
VLM-3R-7B~\cite{fan2025vlm} & \cellcolor{best}{31.4} & \cellcolor{best}{32.9} & \cellcolor{second}{44.4} & \cellcolor{best}{40.8} & \cellcolor{best}{44.0} & \cellcolor{best}{37.7} & \cellcolor{best}{38.7} \\
\midrule
\rowcolor{blue!5}\multicolumn{8}{l}{\textbf{Open-Source General Models}} \\
Qwen2-VL-2B-Instruct~\cite{wang2024qwen2vl} & 29.0 & 36.1 & 45.6 & 44.1 & 39.2 & 37.8 & 38.8 \\
Qwen2-VL-7B-Instruct~\cite{wang2024qwen2vl} & \cellcolor{best}{31.4} & \cellcolor{second}{40.2} & \cellcolor{best}\underline{49.8} & \cellcolor{second}{49.5} & \cellcolor{best}{45.3} & \cellcolor{best}{42.0} & \cellcolor{best}{43.2} \\
Qwen2.5-VL-3B-Instruct~\cite{bai2025qwen25vltechnicalreport} & \cellcolor{second}{31.3} & 40.1 & 46.7 & 45.6 & \cellcolor{second}{43.4} & 40.8 & 41.4 \\
Qwen2.5-VL-7B-Instruct~\cite{bai2025qwen25vltechnicalreport} & 30.6 & 32.7 & 47.5 & 49.3 & 42.9 & 38.9 & 40.6 \\
LLaVa-onevision-qwen2-7B~\cite{li2024llava} & 30.0 & 37.0 & 45.6 & 45.9 & 41.9 & 39.0 & 40.1 \\
LLaVA-Video-7B~\cite{zhang2024llavavideo} & 25.6 & 31.3 & 40.5 & 36.5 & 32.7 & 32.6 & 33.3 \\
LLaVA-NeXT-Video-7B~\cite{zhang2024llavanextvideo} & 24.5 & 29.5 & 39.8 & 34.7 & 37.5 & 32.4 & 33.2 \\
InternVL3-2B~\cite{zhu2025internvl3} & 29.2 & 32.9 & 37.9 & 39.7 & 33.6 & 34.0 & 34.7 \\
InternVL3-8B~\cite{zhu2025internvl3} & 23.9 & 34.5 & 46.9 & 44.3 & 40.7 & 36.7 & 38.1 \\
MiniCPM-V-2.6~\cite{hu2024minicpm} & 27.1 & 30.4 & 48.3 & 44.8 & 38.8 & 36.4 & 37.9 \\
Molmo-7B-D~\cite{deitke2024molmopixmoopenweights} & 25.3 & 37.7 & 32.0 & 36.0 & 31.2 & 32.3 & 32.4 \\
mPLUG-Owl3-7B~\cite{ye2024mplugowl3} & 27.5 & 37.7 & 42.2 & 42.9 & 38.3 & 36.9 & 37.7 \\
VILA1.5-7B~\cite{lin2024vila} & 28.1 & 32.5 & 41.3 & 39.7 & 37.5 & 34.9 & 35.8 \\
Ovis2.5-9B~\cite{lu2024ovis} & 27.1 & \cellcolor{best}{41.7} & \cellcolor{second}{49.0} & \cellcolor{best}{53.2} & 42.5 & \cellcolor{second}{41.2} & \cellcolor{second}{42.7} \\
Oryx-7B~\cite{liu2025oryxmllmondemandspatialtemporal} & 24.3 & 30.8 & 32.8 & 32.0 & 29.7 & 29.6 & 29.9 \\
Phi-3.5-Vision-Instruct-4B~\cite{abdin2024phi3technicalreporthighly} & 27.3 & 31.9 & 36.6 & 37.2 & 31.4 & 32.3 & 32.9 \\
\bottomrule
\end{tabular}%
}
\end{table}

\subsection{Evaluation Setup}
We evaluate 23 state-of-the-art MLLMs, spanning 19 open-source models and 4 proprietary models, as listed in Table~\ref{tab:main_results}. The open-source models include both general-purpose and spatial MLLMs. Open-source models are evaluated using their official implementations, while proprietary models are accessed through their APIs. All models are evaluated zero-shot on the multiple-choice VQA task. We report Overall accuracy, computed as the total number of correct predictions over all questions, and Average accuracy, computed as the mean of task-wise accuracies across the five tasks.

\subsection{Results}
\Cref{tab:main_results} reveals a substantial gap between human and model performance, highlighting limitations of current MLLMs in collaborative multi-agent reasoning. Proprietary models outperform open-source ones, with Gemini-2.5-Pro achieving the highest accuracy, while the strongest open-source model, Ovis2.5-9B, trails by about 14\%, and most others cluster between 30–42\%.

While models struggle across all reasoning categories, the largest difficulties appear in tasks requiring compositional reasoning over geometric transformations (\textit{i.e.}, Navigation and Egocentric Motion), where many models approach near-random accuracy. In contrast, models perform relatively better on Occlusion and Relative Position, indicating stronger capability in coarse spatial perception than in dynamic, viewpoint-dependent geometric reasoning. Overall, these results suggest that current MLLMs rely more on orientation-specific cues than on robust geometric reasoning for collaborative multi-agent reasoning.

\section{Analysis and Discussion}
\label{sec:exp}
This section analyzes limitations in multi-agent reasoning, focusing on gaps between spatial perception and geometry, increasing relational complexity, and geometric consistency.

\subsection{From Spatial Perception to Geometric Reasoning}
\label{subsec:4.1}

\begin{figure}[t!]
    \centering

    \includegraphics[width=1.0\linewidth]{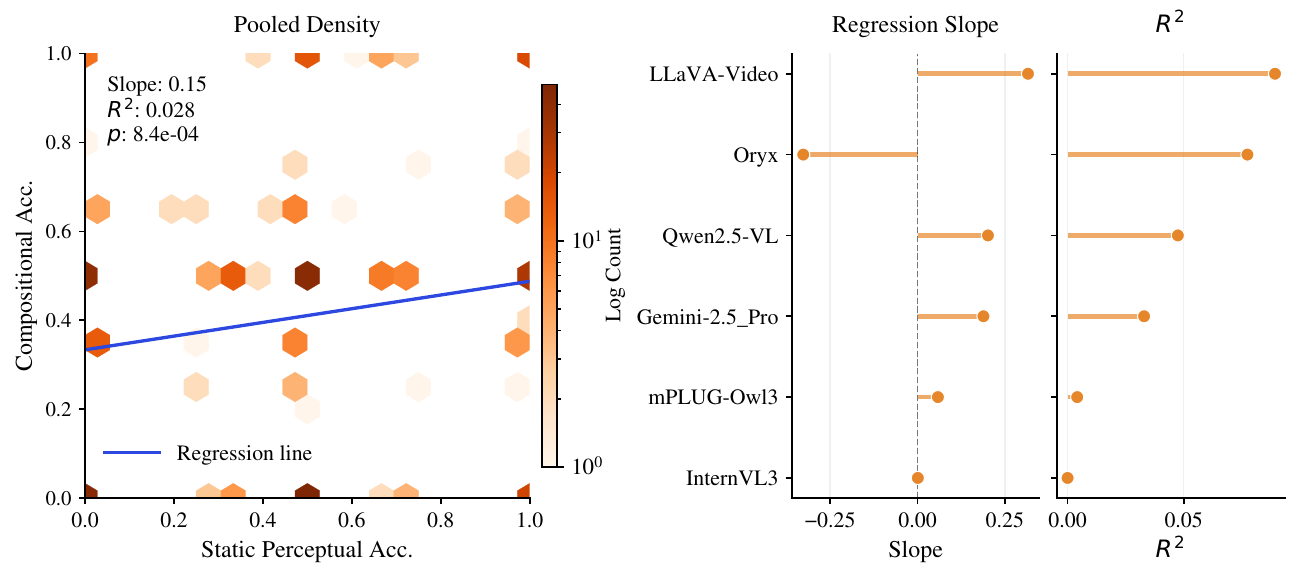}
    \caption{\textbf{Relationship between spatial perception and geometric reasoning.}
    \textit{Left}: Hexbin density of static perception vs. compositional accuracy pooled across models, with linear regression fit.
    \textit{Right}: Per-model regressions of compositional accuracy on static perception accuracy, with \textit{right}$_1$ showing regression slope (geometric scaling strength), and \textit{right}$_2$ reporting regression $R^2$ values (compositional variance over accuracy explained by static perception accuracy).}
    \label{fig:perception_vs_geometric}
\end{figure}

\noindent \textbf{Relationship between Static Perception and Compositional Reasoning over Geometric Transformations.}
To investigate the performance gap between static perception tasks and compositional tasks observed in \Cref{tab:main_results}, we analyze whether strong static spatial perception implies heightened ability in sequencing geometric transformations.
Upon examination, we find weak relationship between model performance on compositional and static perception tasks (\Cref{fig:perception_vs_geometric}).
Although the pooled results show a slight positive trend, per-model regressions yield consistently low $R^2$ values, indicating that degradation in geometric reasoning performance is insufficiently explained by spatial perception. This suggests that while models can reconcile static spatial relationships from multiple egocentric observations, this does not  imply the ability to reliably resolve sequences of geometric transformations.

\begin{figure}
    \centering
    \includegraphics[width=1.0\linewidth]{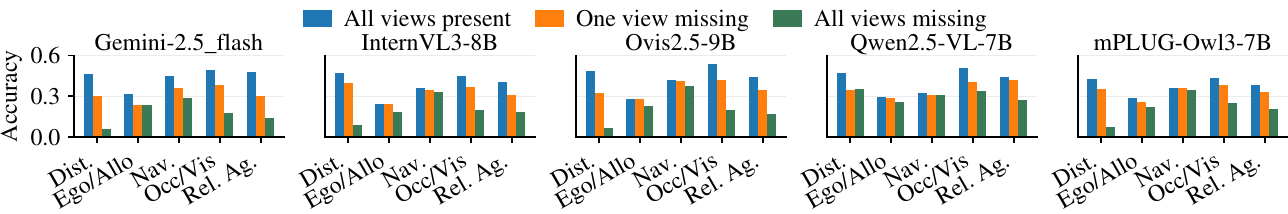}
    \caption{\textbf{Impact of removal of observation on reasoning performance.} Removing agents' observations simulates loss of shared visual context. 
    }
    \label{fig:removed_viewpoints}
\end{figure}

\noindent \textbf{Effect of Removal of Observations.}

To further probe the source of this gap, we analyze whether models integrate information across multiple viewpoints
through a viewpoint removal ablation where one randomly selected observation or all observations are withheld while the question remains unchanged.
Results reveal a clear contrast between proprietary and open-source models (\Cref{fig:removed_viewpoints}). Proprietary models exhibit consistent performance degradation across reasoning categories when viewpoint(s) is removed, indicating that their predictions depend on information distributed across multiple observations. In contrast, open-source models show only minor performance changes, with some categories exhibiting negligible degradation. Notably, in compositional tasks involving geometric primitives, open-source models often maintain above or near random accuracy even with no visual input, suggesting reliance on answer priors or bias rather than cross-agent visual evidence.

\textbf{Takeaway:} Strong static perception does not translate into reliable geometric reasoning in MLLMs because they fail to integrate distributed multi-agent collaborative observations into single unified representations.

\subsection{Cross-Agent Correspondence under Relational Complexity}
\label{subsec:4.2}
\begin{figure}[t!]
    \centering
    \includegraphics[width=\textwidth]{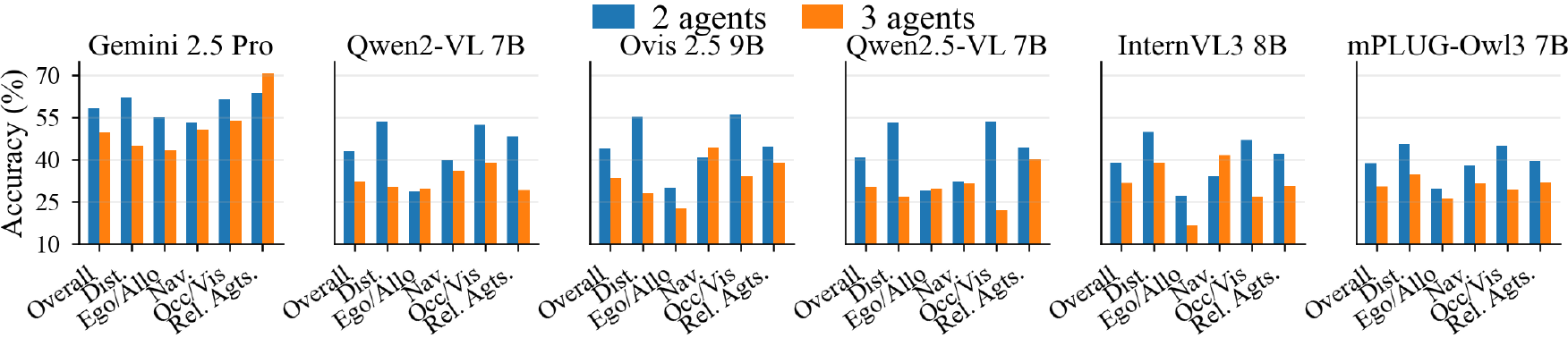}
    \caption{\textbf{Impact of increasing relational complexity.} Performance degrades from two to three agents, indicating weaker cross-agent alignment under higher relational complexity.
    }
    \label{fig:multi_agent_scaling}
\end{figure}

\noindent \textbf{Impact of Increasing Relational Complexity.}
To analyze how relational complexity affects multi-agent collaborative reasoning, we compare performance between two-agent and three-agent observations. This setup directly evaluates the models’ ability to maintain consistent spatial alignment as the number of cross-agent relationships increases. This increase in complexity arises because models must integrate information from more agent viewpoints and resolve a larger set of cross-agent correspondences simultaneously. As shown in \Cref{fig:multi_agent_scaling}, accuracy declines across nearly all models and reasoning categories when transitioning from two-agent to three-agent scenes. The largest drops occur in Navigation and Egocentric Motion, which already exhibit lower baseline performance. These tasks require maintaining spatial relationships across viewpoints, and the additional agent introduces more potential cross-view correspondences that must be resolved simultaneously. The consistent degradation indicates that many MLLMs struggle to maintain coherent cross-agent spatial alignment as relational complexity increases, particularly in geometric tasks.

\noindent \textbf{Effect of Shared Environmental Landmarks.} To further examine the source of this difficulty, we analyze how the density of shared environmental landmarks affects reasoning. Anchors are static environmental elements visible across multiple agents' observation that can serve as reference points for cross-agent alignment. We group observations by anchor count ([0–2], [3–5], [6–9], [10+]) and report results averaged across both two-agent and three-agent observation settings. Results show a non-monotonic pattern (Figure~\ref{fig:anchor_density} in the Supplementary): accuracy improves as anchor density increases from very sparse scenes, but degrades in more cluttered environments. Navigation and Egocentric Motion tasks are particularly sensitive to this effect. While a few shared landmarks provide useful reference points for cross-view alignment, higher landmark density introduces multiple plausible correspondences across viewpoints, increasing ambiguity in matching spatial references across agents and leading to performance degradation. These results suggest that as relational structure grows, models struggle to consistently resolve cross-agent correspondence.

\textbf{Takeaway:} As independent observations from multiple agents and scene clutter increase, MLLMs struggle to resolve cross-view correspondences, leading to unstable reasoning.

\subsection{Geometric Consistency and Structural Cues}
\label{subsec:4.3}

\begin{figure*}[t]
    \centering
    \includegraphics[width=0.94\textwidth]{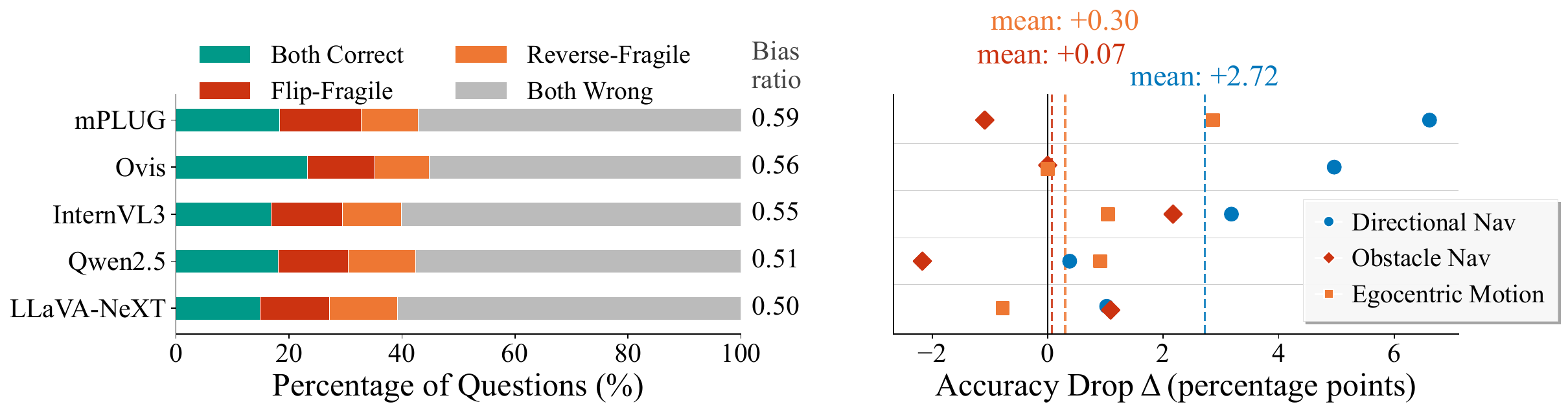}
    \caption{\textbf{Orientation consistency analysis under horizontal flipping.} \textit{Left:} Question outcomes classified by correctness across original and flipped orientations. Here, Flip-fragile: correct only in original; reverse-fragile: correct only when flipped. Bias ratio\,=\,flip-fragile\,/\,(flip-fragile\,+\,reverse-fragile); 0.50 indicates symmetric fragility. \textit{Right} Accuracy change under reflection by reasoning category; dashed lines denote mean shifts across models.
    }
    \label{fig:flip_sensitivity}
\end{figure*}

\noindent \textbf{Orientation Consistency under Symmetry-Preserving Transformation.}
Following the observed degradation under increasing relational complexity, we examine whether these failures stem from unstable spatial representations. We hypothesize that if models maintain stable spatial representations, their predictions should remain invariant under symmetry-preserving transformations; failure indicates orientation-sensitive reasoning. To assess this hypothesis, we perform a horizontal reflection test on Navigation and Egocentric Motion tasks, where images are flipped and directional language is correspondingly swapped (e.g., “turn left” $\rightarrow$ “turn right”) to preserve geometric equivalence. Despite this symmetry, many predictions change after reflection, and only a minority remain correct in both orientations (\Cref{fig:flip_sensitivity}, \textit{left}), indicating orientation instability under geometrically equivalent transformations.
To further probe the source of this instability, we divide Navigation into directional questions, where both image and text are transformed, and obstacle-based questions, where only the image is flipped. Directional questions exhibit a larger accuracy drop than obstacle-based ones, while Egocentric Motion shows a moderate decline (\Cref{fig:flip_sensitivity}, \textit{right}). This suggests that orientation sensitivity is more influenced by directional language cues than stable geometric representations.

\begin{table}[!b]
\centering
\caption{Human error analysis. \textbf{(a)} Audit of incorrect human responses across 21 questions. \textbf{(b)} Cross-tabulation of human and Gemini-2.5-Pro correctness on the same 81 questions.}
\label{tab:human_analysis}
\setlength{\tabcolsep}{4pt}
\renewcommand{\arraystretch}{0.95}
\small
\begin{subtable}[b]{0.52\linewidth}
    \centering
    \resizebox{\linewidth}{!}{
    \begin{tabular}{@{}lc@{}}
        \toprule
        \textbf{Type of Incorrect Response} & \textbf{Count} \\
        \midrule
        Rotated, mirrored, reversed, or reordered direction  & 13 \\
        Miscounted, misidentified, or misjudged proximity    & 7 \\
        Missed the ``none / not visible'' option             & 2 \\
        Selected ``neither can reach'' despite a valid route & 1 \\
        \midrule
        \textbf{Total} & \textbf{23} \\
        \bottomrule
    \end{tabular}
    }
    \caption{}
    \label{tab:human_errors}
\end{subtable}\hfill
\begin{subtable}[b]{0.44\linewidth}
    \centering
    \resizebox{\linewidth}{!}{
    \begin{tabular}{@{}lccc@{}}
        \toprule
         & \textbf{Gem.\ Corr.} & \textbf{Gem.\ Incorr.} & \textbf{Total} \\
        \midrule
        All eval.\ correct       & 41 & 19 & 60 \\
        $\geq$1 eval.\ incorrect & 11 & 10 & 21 \\
        \midrule
        \textbf{Total} & \textbf{52} & \textbf{29} & \textbf{81} \\
        \bottomrule
    \end{tabular}
    }
    \caption{}
    \label{tab:human_gemini}
\end{subtable}
\end{table}

\begin{figure*}[t]
    \centering
    \includegraphics[width=\textwidth]{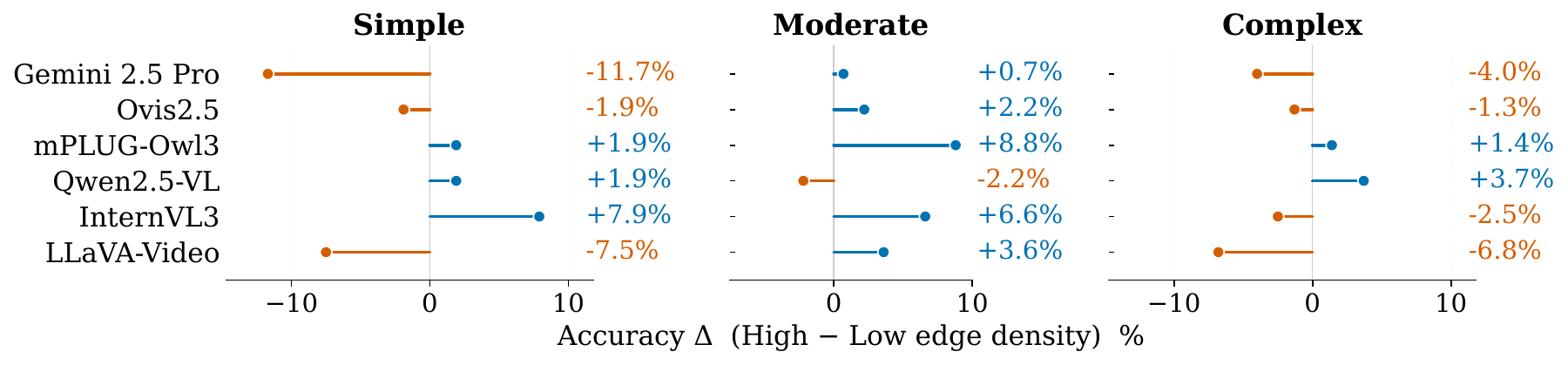}
    \caption{\textbf{Impact of depth cue complexity.} \textbf{(a)} Accuracy change ($\Delta$) between high and low depth edge density in distinct depth layers.
    }
    \label{fig:depth_cue_analysis}
\end{figure*}

\noindent \textbf{Impact of Depth Cue Complexity.} We next examine whether models can leverage geometric scene structure when it is naturally present in the image by analyzing whether a higher density of depth discontinuities, one type of monocular depth cue, correlates with improved multi-agent reasoning. Relative depth maps are extracted using Depth Anything V2 \cite{yang2024depth}, and depth edge density is computed as the fraction of pixels lying on detected depth boundaries.
Importantly, models are evaluated only with the original RGB images; depth edges are computed post-hoc to analyze whether scenes with richer geometric structure correspond to performance change. To control for scene complexity, samples are grouped by the number of distinct depth layers (Simple: 1–2, Moderate: 3, Complex: 4+), and performance is compared between high- and low-edge-density scenes within each group. As shown in \Cref{fig:depth_cue_analysis}, stronger depth boundary structure does not improve performance; most models perform worse at higher edge density, with the largest degradation in simple scenes, while effects remain small or negligible in moderate and complex scenes. This suggests that models fail to exploit geometric scene structure and often treat additional depth cues as visual complexity rather than useful spatial information.

\textbf{Takeaway:} MLLMs exhibit weak use of geometric cues, with predictions sensitive to orientation changes and largely unaffected by scene geometry, leading to inconsistent reasoning and poor cross-agent alignment.

\subsection{Human Reasoning Difficulty}
\textbf{Types of Human Errors.} Human evaluators reach 73.2\% overall accuracy on MaRiO (Table~\ref{tab:main_results}), with the lowest scores on Egocentric Motion and Navigation. To understand the nature of this difficulty, we examine every question answered incorrectly by at least one evaluator (21 of 81 questions) and categorize 23 incorrect responses by error type (Table~\ref{tab:human_errors}). The dominant error type (13 of 23) is direction confusion, where the evaluator rotated, mirrored, reversed, or reordered the correct direction, consistent with spatial-transformation mistakes. This is expected for three-agent Egocentric Motion questions, which require composing multiple sequential viewpoint transformations and place substantial cognitive demand on the evaluator. Prior work in cognitive science shows that reasoning about a different viewpoint involves a mental-rotation-like process whose time and effort scale with the degree of perspective change, and that judgments become more error-prone as the number of required transformations increases. Visibility-type judgments, which do not require such transformations, do not show the same cost~\cite{michelon2006two, kessler2010embodied}.

\textbf{Human-Model Comparison.} We further compare human and Gemini-2.5-Pro outcomes on the same 81 questions (Table~\ref{tab:human_gemini}). Among the 60 questions every evaluator answered correctly, Gemini-2.5-Pro failed on 19, indicating that human success does not simply track the questions a strong model also answers correctly. Conversely, among the 21 questions with at least one human error, Gemini-2.5-Pro answered 11 correctly. Human and model errors therefore fall on largely different subsets of questions, rather than on a shared set of questions that are difficult for any answerer.

\textbf{Takeaway:} Human errors reflect the cognitive cost of composing viewpoint transformations and do not coincide with model errors, indicating that MaRiO's questions are answerable while remaining demanding for both humans and MLLMs.

\section{View Augmentation for Collaborative Reasoning}

\begin{table}[t!]
\setlength{\tabcolsep}{4pt} 
\begin{minipage}{0.62\columnwidth}
    \centering
    \caption{\textbf{Performance of synthetic view augmentations.} \colorbox{improved}{Improvements} after adding synthetic views and \textbf{\textcolor{darkgreen}{best}} overall gain are highlighted.}
    \resizebox{\linewidth}{!}{
    \begin{tabular}{l|cc|ccc|c|c}
        \toprule
        \multirow{2}{*}{\textbf{Model}} & \multicolumn{2}{c|}{\textbf{Comp.}} & \multicolumn{3}{c|}{\textbf{Static Perc.}} & \multirow{2}{*}{\textbf{Overall}} & \multirow{2}{*}{\textbf{Avg.}}\\
        & \textbf{Ego.} & \textbf{Nav.} & \textbf{Dist.} & \textbf{Occ.} & \textbf{Rel.} & & \\
        \midrule
        \rowcolor{blue!5}\multicolumn{8}{l}{\textbf{Spatial Models}} \\
        VST-7B-RL & 28.3 & 32.8 & 41.6 & 38.9 & 38.3 & 35.1 & 36.0\\
        \hspace{1em}$+$ Syn. View & \cellcolor{improved}29.0 & \cellcolor{improved}35.4 & 25.0 & 36.1 & \cellcolor{improved}41.5 & 33.2 & 33.4\\
        & \textcolor{darkgreen}{\scriptsize\textbf{+0.7}} & \textcolor{darkgreen}{\scriptsize\textbf{+2.6}} & \textcolor{red}{\scriptsize-16.6} & \textcolor{red}{\scriptsize-2.8} & \textcolor{darkgreen}{\scriptsize+3.2} & \textcolor{red}{\scriptsize-1.9} & \textcolor{red}{\scriptsize-2.6}\\
        \midrule
        VLM-3R-7B & 31.4 & 32.9 & 44.4 & 40.8 & 44.0 & 37.7 & 38.7 \\
        \hspace{1em}$+$ Syn. View & \cellcolor{improved}31.9 & 32.6 & 43.8 & 39.4 & \cellcolor{improved}46.0 & \cellcolor{improved}37.8 & 38.7\\
        & \textcolor{darkgreen}{\scriptsize+0.5} & \textcolor{red}{\scriptsize-0.3} & \textcolor{red}{\scriptsize-0.6} & \textcolor{red}{\scriptsize-1.4} & \textcolor{darkgreen}{\scriptsize+2.0} & \textcolor{darkgreen}{\scriptsize+0.1} & \scriptsize0.0\\
        \midrule
        \rowcolor{blue!5}\multicolumn{8}{l}{\textbf{General Models}} \\
        LLaVA-Video-7B & 25.6 & 31.3 & 40.5 & 36.5 & 32.7 & 32.6 & 33.3 \\
        \hspace{1em}$+$ Syn. View & 24.7 & \cellcolor{improved}32.9 & 40.2 & 33.6 & \cellcolor{improved}35.6 & \cellcolor{improved}32.9 & \cellcolor{improved}33.4\\
        & \textcolor{red}{\scriptsize-0.9} & \textcolor{darkgreen}{\scriptsize+1.6} & \textcolor{red}{\scriptsize-0.3} & \textcolor{red}{\scriptsize-2.9} & \textcolor{darkgreen}{\scriptsize+2.9} & \textcolor{darkgreen}{\scriptsize+0.3} & \textcolor{darkgreen}{\scriptsize+0.1}\\
        \midrule
        Qwen2.5-VL-7B & 30.6 & 32.7 & 47.5 & 49.3 & 42.9 & 38.9 & 40.6 \\
        \hspace{1em}$+$ Syn. View & 30.6 & \cellcolor{improved}33.3 & 45.2 & \cellcolor{improved}53.8 & \cellcolor{improved}46.8 & \cellcolor{improved}40.0 & \cellcolor{improved}41.9\\
        & \scriptsize0.0 & \textcolor{darkgreen}{\scriptsize+0.6} & \textcolor{red}{\scriptsize-2.3} & \textcolor{darkgreen}{\scriptsize\textbf{+4.5}} & \textcolor{darkgreen}{\scriptsize\textbf{+3.9}} & \textcolor{darkgreen}{\scriptsize\textbf{+1.1}} & \textcolor{darkgreen}{\scriptsize\textbf{+1.3}}\\
        \bottomrule
    \end{tabular}}
    \label{tab:main_app}
\end{minipage}
\hfill
\begin{minipage}{0.36\columnwidth}
    \centering
    \includegraphics[width=\linewidth]{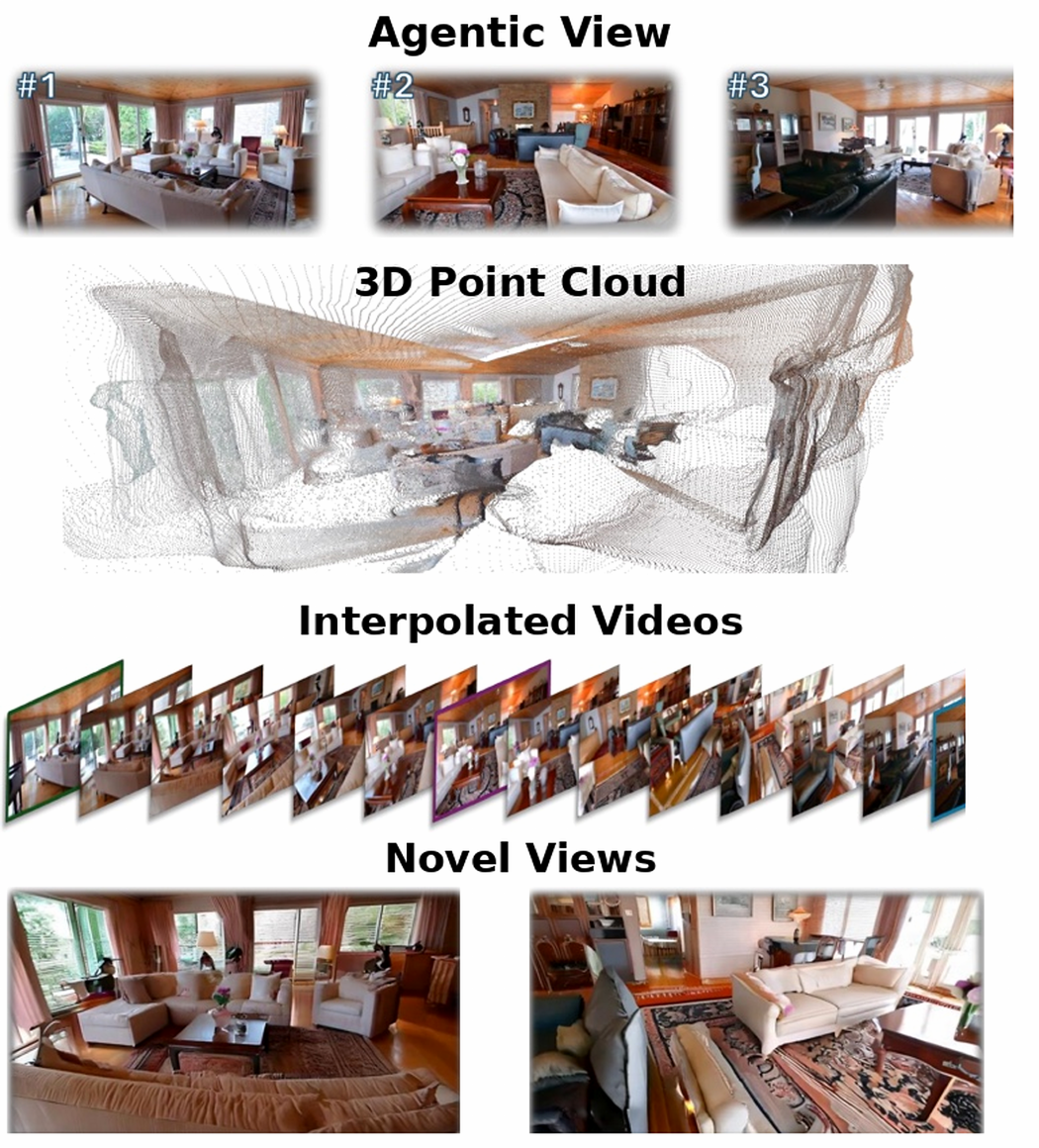}
    \captionof{figure}{\textbf{Geometry guided synthetic view augmentation.} Synthetic transitions interpolate between agent viewpoints to produce intermediate views, providing auxiliary spatial context between observations.}
    \label{fig:syn_vids}
\end{minipage}
\end{table}

Our analysis reveals that many failures stem from models being unable to reliably establish correspondences across agents’ independent observations, as these tasks require jointly reasoning over all observations to reach a single decision. To mitigate this, we test whether providing additional intermediate viewpoints between observations can help bridge the spatial gap across agents. Intuitively, if transitions between viewpoints are available, they can reveal how observations relate spatially, allowing the model to use these frames as auxiliary visual context rather than inferring cross-view relationships purely through reasoning over independent observations.

\subsection{Geometry-Guided Synthetic View Augmentation}
We generate synthetic intermediate viewpoints to bridge the spatial gap between agents’ observations and provide additional context. First, we reconstruct a 3D point-cloud representation \cite{wang2024dust3r} of the scene from the multiple agents' observation. Camera poses are then interpolated and intermediate views are rendered from the reconstructed geometry to approximate the spatial transitions between observations. These rendered views are used to guide a diffusion-based video generation process \cite{yu2024viewcraftertamingvideodiffusion} that produces a sequence of synthetic frames containing novel intermediate viewpoints. The resulting video provides additional visual context that reveals how the independent observations relate spatially. For models that support video input, the generated video is provided alongside the original images; for image-only models, sampled frames from the generated sequence are appended to the input images, with the textual prompt modified accordingly. \Cref{fig:syn_vids} shows examples from each stage of the geometry-guided view synthesis process, from reconstructed point clouds to interpolated videos and the resulting novel views.

\noindent \textbf{Effect of Synthetic View Augmentation.} Synthetic view augmentation yields category-specific improvements rather than uniform gains (\Cref{tab:main_app}). Spatial reasoning models show limited benefit from the generated views, with performance remaining unchanged or degrading slightly in some cases. These models already attempt to infer geometric structure from the input images, and diffusion-generated intermediate views may introduce small geometric inconsistencies that affect precise metric reasoning. In contrast, general MLLMs benefit more noticeably from synthetic views, particularly in Navigation and Relative Position tasks, which require aligning spatial relationships across viewpoints. The generated transitions likely provide useful intermediate cues that help models connect observations from different agents. Ablations on the number of sampled frames, interpolation order, and a cropped-frame baseline confirm that gains arise from the synthesized novel viewpoints rather than from additional frames alone (see Supplementary).

\textbf{\textit{Takeaway}:} Synthetic augmentation selectively improves multi-agent reasoning by aiding cross-view relational understanding, while offering limited benefit for precise geometric estimation tasks.

\section{Conclusion}
\label{sec:conclusion}

In this work, we introduced MaRiO, a benchmark for evaluating multi-agent collaborative reasoning, and studied the behavior of 23 state-of-the-art open-source and proprietary MLLMs on this task. Our analysis reveals a substantial gap between human and model performance, with models struggling particularly on geometric tasks such as Navigation and Egocentric Motion. These failures largely stem from difficulty in establishing reliable cross-agent correspondences across independent viewpoints. As an initial step toward addressing these challenges, we explore geometry-guided synthetic view augmentation that introduces intermediate viewpoints between observations, providing additional spatial context that improves performance in certain reasoning categories. Overall, our results highlight the need for stronger cross-agent alignment and geometry-aware perception in future multimodal embodied systems for collaborative environments.

\paragraph{Limitations.}
MaRiO focuses on static, snapshot-based observations and a multiple-choice QA setup, which does not capture temporal dynamics or interactive decision-making. Scores in this format can also depend on the number of answer options and do not directly measure a model's ability to answer without them. Extending the benchmark to sequential observations and embodied interactions would provide a fuller evaluation of multi-agent reasoning in realistic settings.

\paragraph{Broader Impact.}
Our findings highlight gaps in current MLLMs for multi-agent spatial reasoning, with implications for applications such as robotics, autonomous navigation, and collaborative augmented reality, where reliable cross-agent understanding is essential. However, improving such capabilities raises considerations around deployment in safety-critical and privacy-sensitive environments, including multi-camera surveillance, emphasizing the need for robust, interpretable, and responsible use.

\begin{ack}
This work received no third-party funding or support. The authors declare no competing interests.
\end{ack}

\bibliographystyle{ieeetr}
\bibliography{neurips_2026}

\newpage

\newpage
\appendix

{\centering
{\large \textbf{
MaRiO: Multi-agent Collaborative Reasoning via Shared Observations in MLLMs\\ \vspace{1mm}}{ \textit{\textbf{Supplementary Material}}}}
\par}

\vspace{2mm}
\renewcommand{\thesection}{\Alph{section}}
\renewcommand{\thesubsection}{\thesection.\arabic{subsection}}
\setcounter{section}{0}

\noindent In this supplementary material, we present additional analyses and implementation details supporting the main paper. Specifically, we include model bias analysis (\Cref{sec:bias-supp}), and generalization of our findings to additional models (\Cref{sec:gen-supp}). Then, we provide details of the benchmark construction and its validation (\Cref{sec:bench-supp}), and qualitative examples illustrating model behaviors (\Cref{sec:qual-supp}). Lastly, we conclude with a discussion of future research directions (\Cref{sec:future}).

\etocdepthtag.toc{mtappendix}
\etocsettagdepth{mtchapter}{none}
\etocsettagdepth{mtappendix}{subsection}
\tableofcontents

\newpage

\section{Model Bias Analysis}
\label{sec:bias-supp}
To better understand the internal reasoning tendencies of MLLMs, we conduct a series of diagnostic analyses aimed at identifying systematic biases in spatial reasoning behavior. Rather than evaluating overall accuracy, these analyses probe how models respond to different structural formulations of spatial tasks. Our findings reveal consistent biases in how models represent navigation primitives and directional semantics.

\noindent \textbf{Bias Towards Navigational Primitives.} To analyze how models reason over and utilize navigation primitives, we evaluate performance across subsets of navigation questions that isolate translation and rotation motion primitives. If models internally represent navigation as a sequence of grounded spatial operations, their performance should remain relatively stable across primitive types. Conversely, systematic differences may reveal biases in how models interpret or reason about movement instructions. 

\begin{figure}[t!]
    \centering
    \includegraphics[width=\linewidth]{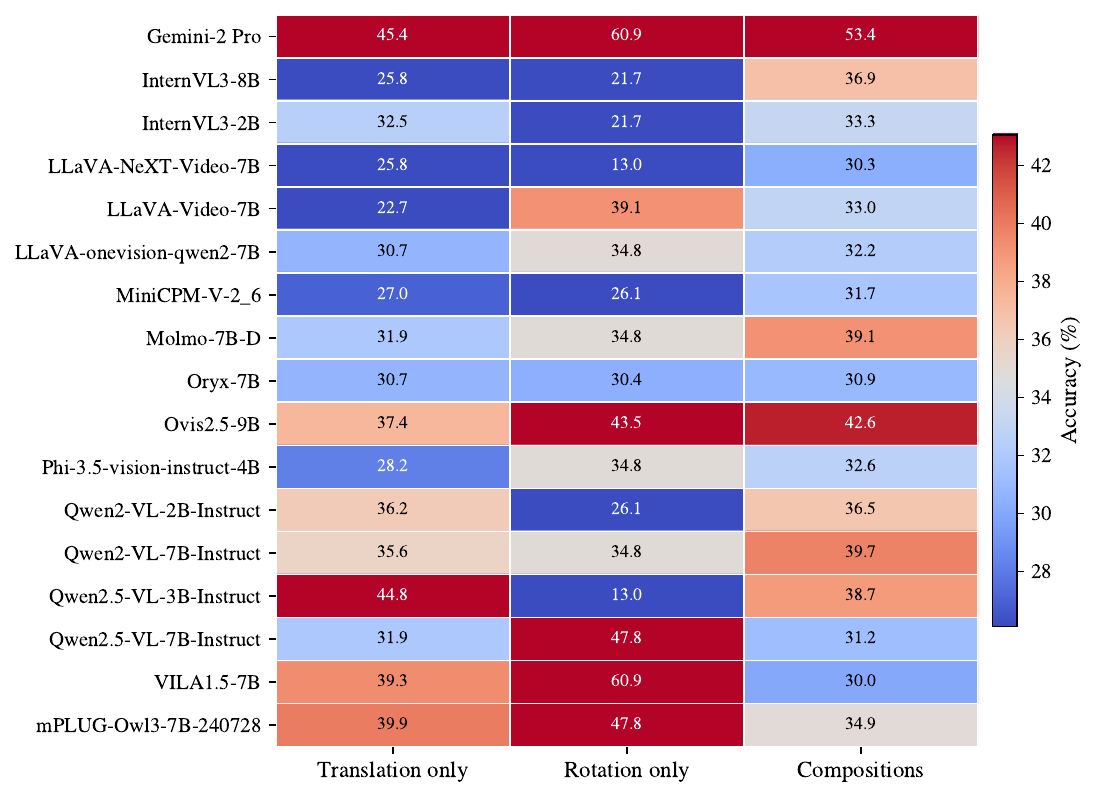}
    \caption{
    \textbf{Bias towards navigational primitives.} Heatmap displaying MLLM accuracy for navigation QA subsets where 'Translation only' refers to navigation QA samples whose answer choices consist only of translation primitives, 'Rotation only' refers to navigation QA samples whose answer choices consist only of rotation primitives, and 'Compositions' refers to navigation QA samples that have at least one answer choice with a combination of translation and rotation primitives
    }
    \label{fig:nav_heatmap}
\end{figure}

We partition the navigation QA subset according to the primitives present in answer choices: \textit{Translation Only}, \textit{Rotation Only}, and \textit{Compositions}, where the latter contains action sequences consisting of both translation and rotation. For each model, we compute Top-1 Accuracy separately for each of these subsets. 
As shown in \Cref{fig:nav_heatmap}, models exhibit clear asymmetries across primitive types. Several models perform substantially better on translation-only questions, while others show higher accuracy for rotation-only scenarios. Interestingly, some models achieve higher accuracy on compositional actions than on either primitive subset, suggesting that structured multi-step instructions may provide contextual cues that facilitate reasoning. In contrast, other models show the opposite behavior, where performance degrades when primitives must be combined.

\textbf{Takeaway:} MLLMs differ in how they represent navigational primitives. While some appear to benefit from compositional structure, others struggle to integrate multiple spatial operations into a coherent navigation plan.

\begin{figure}[t!]
    \centering
    \includegraphics[width=\linewidth]{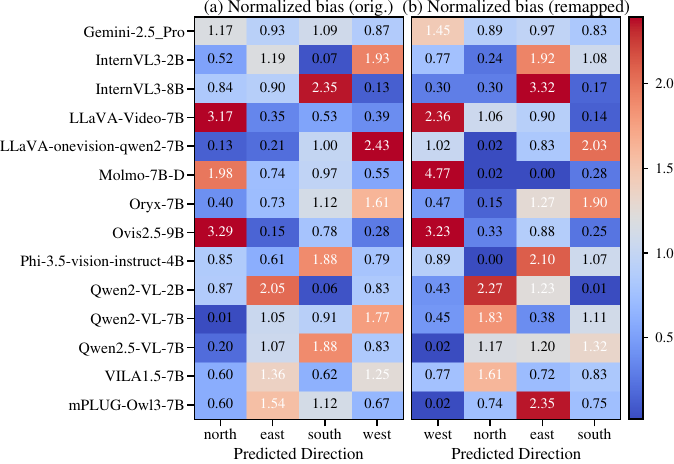}
    \setlength{\abovecaptionskip}{2pt}
    \caption{\textbf{Egocentric directional bias across models.}  
    Predicted cardinal frequencies normalized by the ground-truth prior are shown under the original binding (\textit{left}) and a 90° clockwise remapped binding (\textit{right}). Values $>1$ indicate systematic overprediction relative to task frequency. In the original mapping, \textit{north}, \textit{east}, \textit{south}, and \textit{west} correspond to forward, rightward, backward, and leftward motion, respectively.
    }
    \label{fig:carinal_bias_heatmap}
\end{figure}

\noindent \textbf{Directional Bias under Cardinal Remapping.}
We next examine whether models bind spatial semantics to egocentric spatial frames in a geometry-consistent manner. Specifically, we analyze directional predictions under two binding conditions:
(1) the original binding between cardinal tokens and egocentric spatial relations (forward, left, right, backward), and (2) a remapped binding in which these associations are permuted. For each model, we compute predicted cardinal token frequencies normalized by the ground-truth prior. Results are shown in \Cref{fig:carinal_bias_heatmap}.

Under remapped bindings, models exhibit heterogeneous behavior. Some architectures (e.g., LLaVA-Video-7B, Ovis2.5-9B) display rotation-consistent shifts in their directional predictions (e.g., North→West), suggesting that directional reasoning is partially grounded in an egocentric spatial representation  rather than arbitrary token-level association. In contrast, other models (e.g., InternVL3-8B, mPLUG-Owl3) maintain persistent token-level directional preferences that do not rotate coherently after remapping.

\textbf{Takeaway:} Directional reasoning mechanisms vary across MLLMs, with some exhibiting partial geometric grounding and others demonstrating stronger token-level associations.

\begin{figure}[t!]
    \centering
    \includegraphics[width=0.5\linewidth]{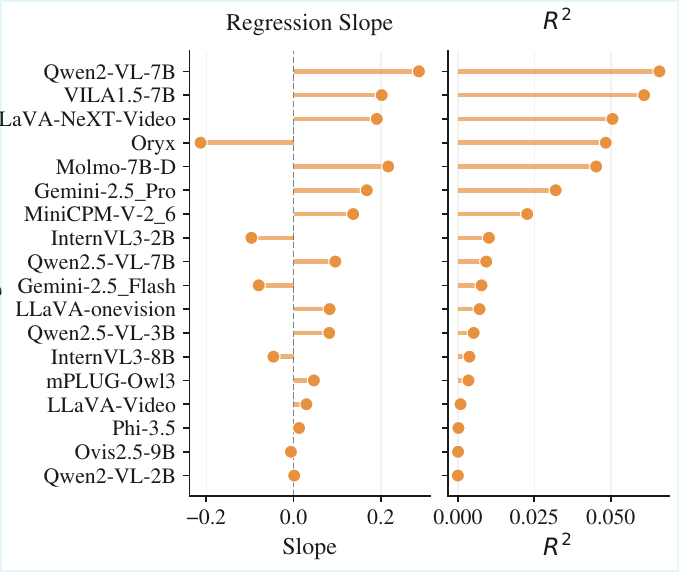}
    \caption{\textbf{Relationship between spatial perception and geometric reasoning.}
    Per-model regressions of geometric accuracy on spatial accuracy, with \textit{left} showing regression slope (geometric scaling strength), and \textit{right} reporting regression $R^2$ values (geometric variance explained by spatial perception).}
    \label{fig:perception_vs_geometric_supp}
\end{figure}

\section{Generalization of Primary Findings}
\label{sec:gen-supp}
In this section, we extend our analysis to a broader set of models to verify whether the findings observed in the main paper (\Cref{sec:exp}) hold across architectures and scales.

\subsection{Relationship between Spatial Perception and Geometric Reasoning}
As shown in \Cref{fig:perception_vs_geometric_supp}, the overall trend remains consistent with our primary findings. While some models exhibit a moderate positive regression slope, indicating that improvements in spatial perception can contribute to better geometric reasoning, the explained variance remains relatively low for most models. This suggests that strong spatial perception alone does not account for geometric reasoning ability. 

\subsection{Effect of Removal of Observations}
\begin{figure}[t!]
    \centering
    \includegraphics[width=1.0\linewidth]{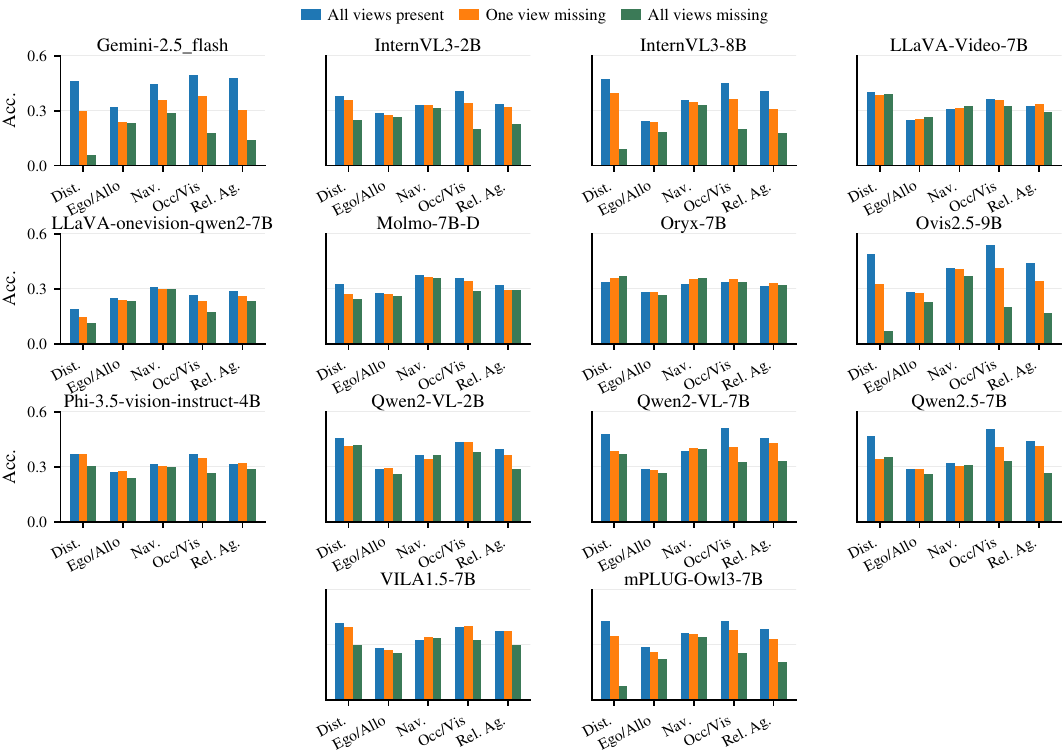}
    \caption{\textbf{Impact of removal of observation on reasoning performance.} Here, removing agents' observations simulates loss of shared visual context. 
    }
    \label{fig:removed_viewpoints_supp}
\end{figure}

As shown in \Cref{fig:removed_viewpoints_supp}, the overall trend remains consistent with our primary findings. Across many open-source models, performance remains noticeably above random even when all views are removed, suggesting that predictions are influenced by dataset biases rather than grounded reasoning over the available observations. However, even when all observations are available, overall gains remain limited, suggesting that many models struggle to effectively integrate distributed multi-agent observations into a coherent spatial representation.

\subsection{Effect of Shared Environmental Landmarks}
\begin{figure}[t]
    \centering
    \includegraphics[width=\textwidth]{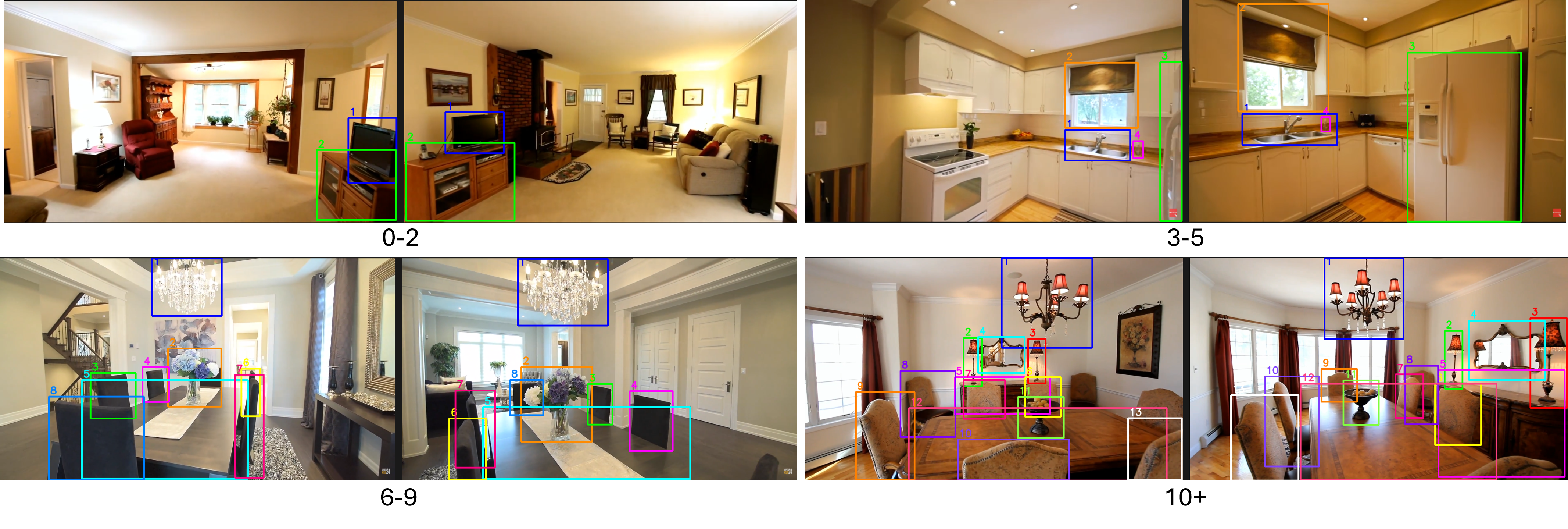}
    \caption{\textbf{Examples of shared anchor density across agent viewpoints}. Each image pair corresponds to two synchronized agent perspectives observing the same environment. Objects visible in both views are marked as shared anchors using consistent color-coded bounding boxes and indices across the pair. Below each pair is their respective anchor-density bin as used in our analysis: $[0,2]$, $[3,5]$, $[6,9]$, and $10+$ shared anchors.}
    \label{fig:shared_anchor_examples}
\end{figure}

\Cref{fig:shared_anchor_examples} illustrates representative examples of the shared anchor density bins used throughout our analysis. Shared anchors correspond to objects that are simultaneously observable across agent viewpoints and thus serve as cross-view reference landmarks that enable spatial reasoning and geometric correspondence across perspectives.

\begin{figure}[t]
    \centering
    \includegraphics[width=0.7\textwidth]{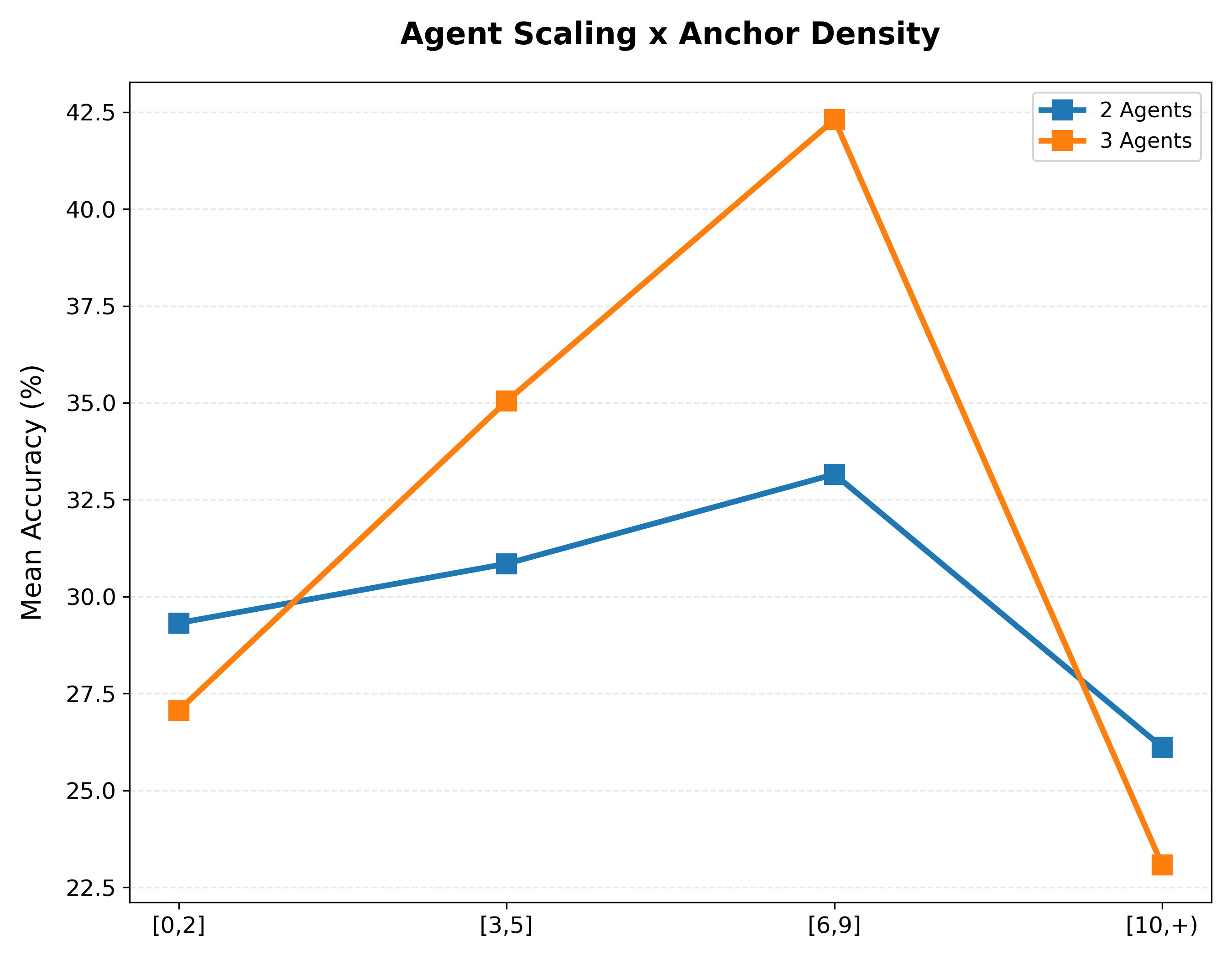}
    \caption{\textbf{Impact of shared environmental landmarks on number of agents}. We observe a performance divergence between 2-agent and 3-agent configurations as anchor density increases. While 3-agent scenarios exhibit a higher peak in moderate density ($[6,9]$ bin), they undergo a sharper degradation under extreme structural complexity.}
    \label{fig:agent_scaling_supp}
\end{figure}

We further analyze performance across different shared anchor densities for both 2-agent and 3-agent settings. \Cref{fig:agent_scaling_supp} shows a  similar trend as our primary finding: performance improves as the number of shared anchors increases up to the ([6,9]) range, suggesting that a moderate number of common landmarks helps models align observations across viewpoints. However, performance degrades in the high-density regime ((10+) anchors), indicating that excessive landmarks introduce clutter and make cross-view correspondence more difficult. This trend appears in both 2-agent and 3-agent configurations, with the effect more pronounced in the 3-agent case, where additional viewpoints increase the relational load required to reconcile spatial correspondences across agents.

\begin{figure}[t!]
  \centering
  \includegraphics[width=0.5\linewidth]{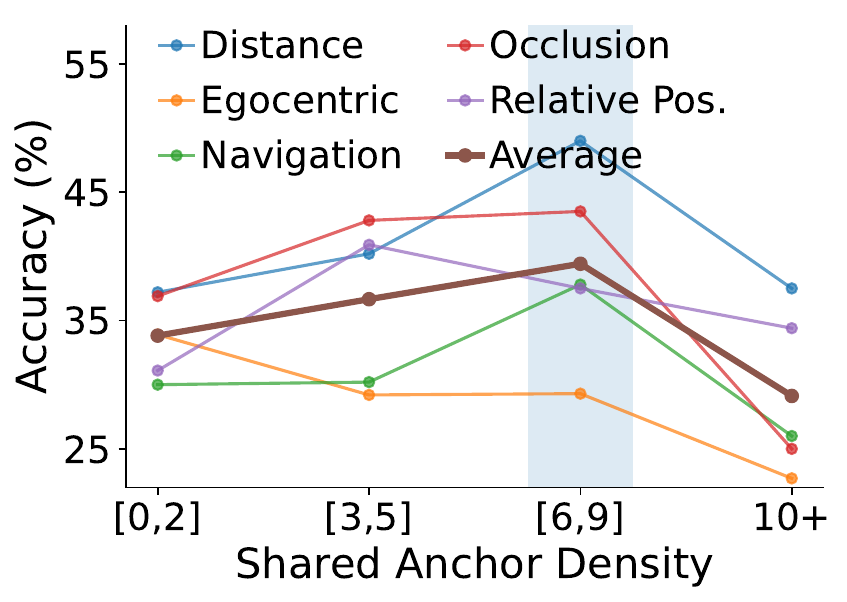}
  \caption{\textbf{Effect of varying density of shared environmental landmarks.}
    Accuracy follows a non-monotonic pattern: it improves as anchor density grows
    from very sparse scenes, then degrades in more cluttered environments.
    Navigation and Egocentric Motion are most sensitive to this effect.}
  \label{fig:anchor_density}
\end{figure}

\subsection{Orientation Consistency Analysis}
\begin{figure}[t!]
    \centering
    \includegraphics[width=\linewidth]{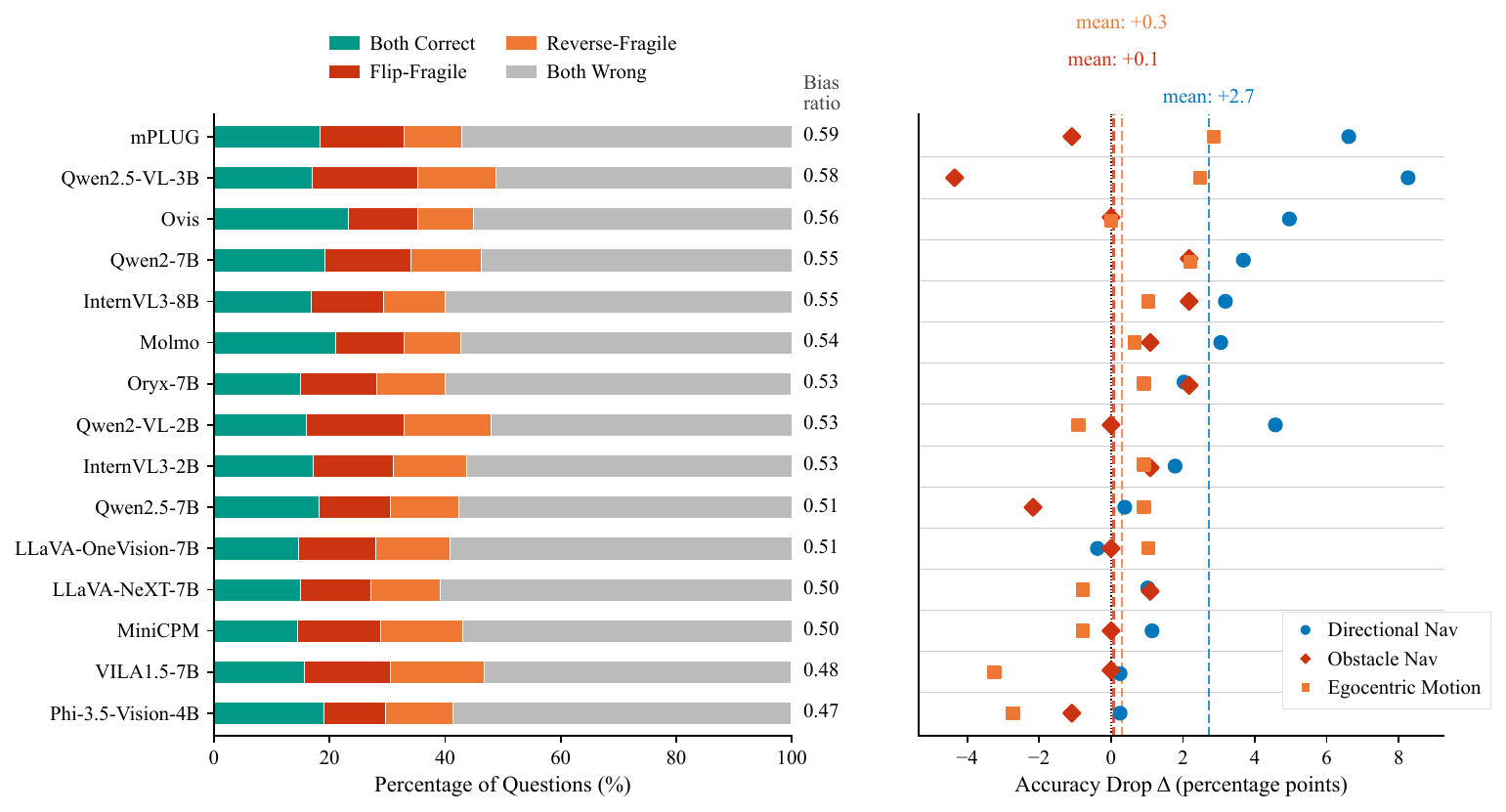}
    \caption{\textbf{Orientation consistency under horizontal reflection for 15 open-source  models.} \emph{Left:} Question outcomes under original and flipped orientations. Bias ratio = flip-fragile\,/\,(flip-fragile + reverse-fragile); 0.50 indicates symmetric fragility. \emph{Right:} Accuracy drop $\Delta$ by reasoning sub-category. Dashed lines indicate cross-model means.}
    \label{fig:supp_flip}
\end{figure}

\Cref{fig:supp_flip} shows the instability observed in the main text generalizes across all architectures: 22-32\% of questions change outcome under horizontal reflection, and bias ratios of 0.47 and 0.59 indicate that none of the model achieves robust orientation invariance. The language-mediated mechanism also holds across models: directional navigation questions show a mean $\Delta$ of ${+2.7}$ percentage points (pp) compared to ${+0.1}$\,pp for obstacle questions where only images change. This finding shows that the instability is driven by directional language cues across  evaluated architectures.

\subsection{Impact of Depth Cue Complexity}
\begin{figure}[t!]
    \centering
    \includegraphics[width=\linewidth]{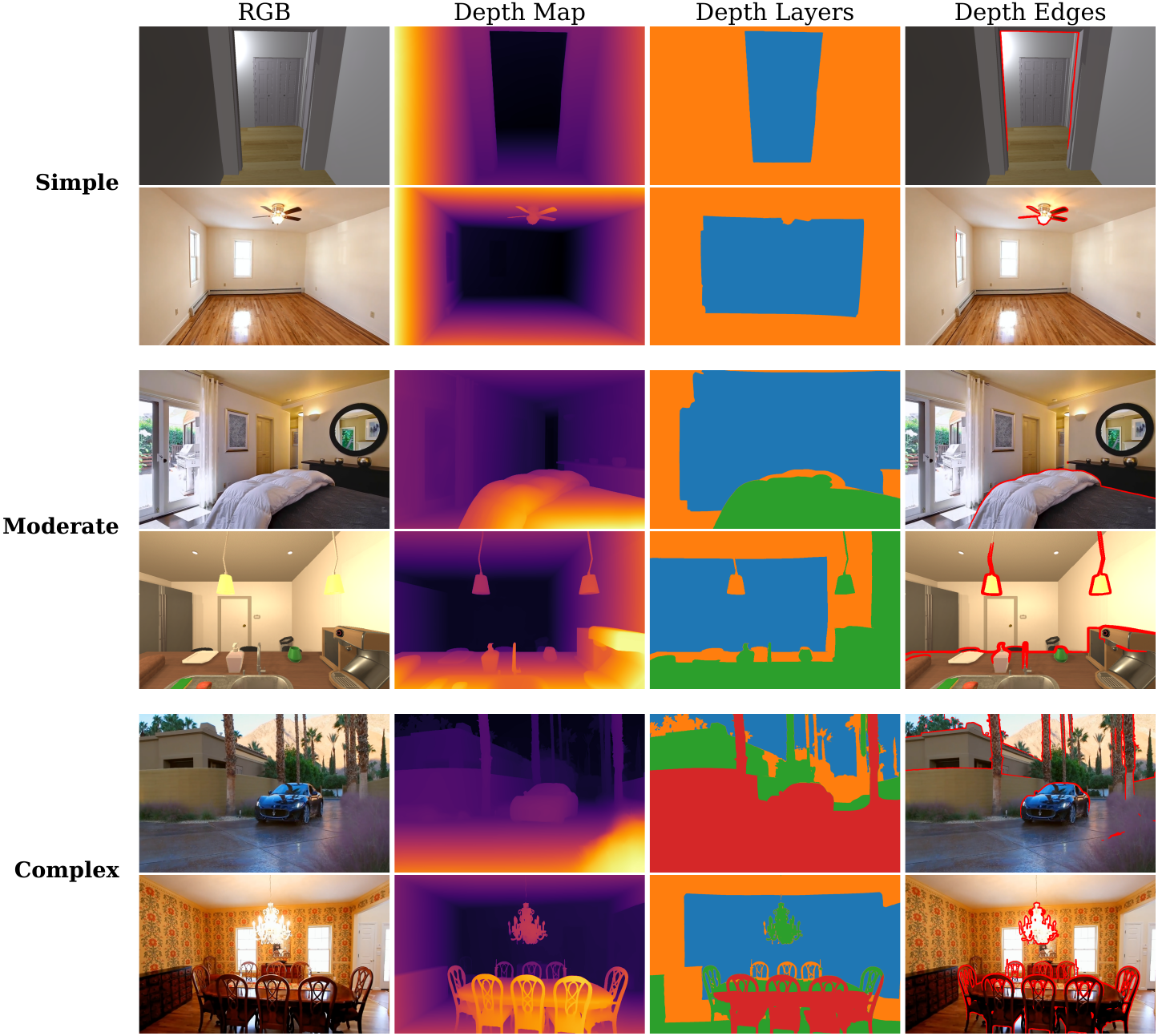}
    \caption{\textbf{Scene complexity across depth layer strata.} Each row shows two examples from a stratum. Columns: original RGB image; monocular depth map; depth layer segmentation by histogram peak assignment (distinct colors indicate distinct depth planes); detected depth edges overlaid in red on the RGB image.}
    \label{fig:supp_depth_strata}
\end{figure}

Figure \ref{fig:supp_depth_strata} illustrates how scene complexity increases across strata, with Simple scenes (1-2 depth layers) containing few structural boundaries and Complex scenes (4+ layers) show extensive depth transitions. Depth edge density is computed on relative depth maps (Depth Anything V2 Large) using Gaussian blur, Sobel gradient magnitude, and Otsu adaptive thresholding. 

\begin{figure}[t!]
    \centering
    \includegraphics[width=\linewidth]{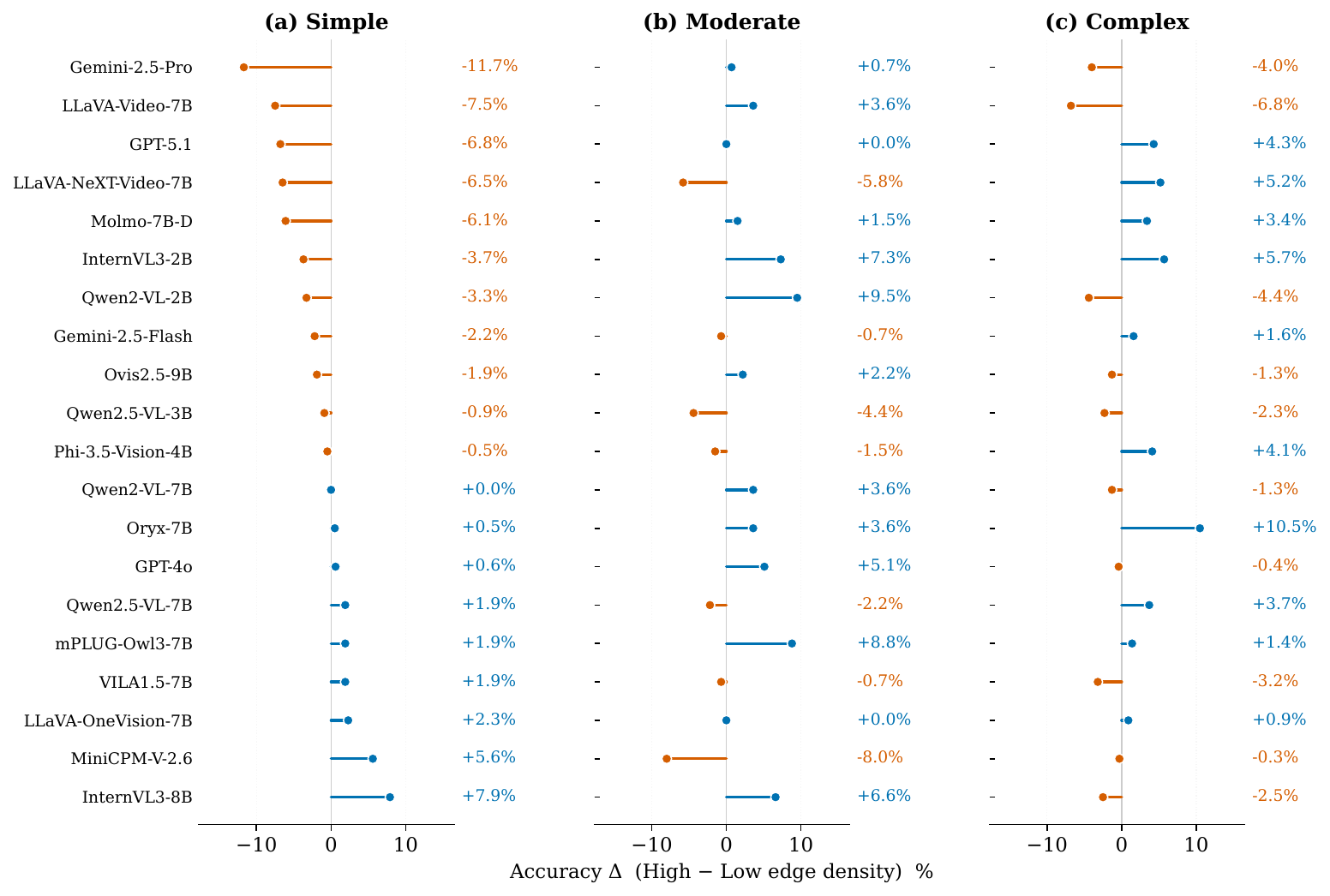}
    \caption{\textbf{Depth edge density effect across 20 models.} Each panel shows the accuracy difference ($\Delta$) between high and low depth edge density samples within a complexity stratum. Only Gemini-2.5-Pro in the Simple stratum shows significant effect after Bonferroni correction ($\alpha = 0.0025$).}
    \label{fig:supp_depth_lollipop}
\end{figure}

The analysis is restricted to the 2-agent subset on Navigation and Egocentric Motion (1221 samples per model). Within each complexity stratum, samples are split at the median edge density and compared using chi-square tests with Bonferroni correction ($\alpha = 0.05/20 = 0.0025$). Figure \ref{fig:supp_depth_lollipop} shows similar trend as our main findings. In the Simple stratum, 11 of 20 models show negative $\Delta$, but only Gemini-2.5-Pro reaches significance ($\Delta = {-11.7\%}$); no model shows a significant positive effect. In the Moderate and Complex strata, the effects are small and inconsistent across models, showing that no evaluated model systematically leverages depth boundary information.

\noindent \textbf{Depth Discontinuities and Scene Clutter.} We focus on depth discontinuities, without covering other monocular cues such as perspective or texture gradients. To check whether depth edge density reflects scene clutter, we compare it with RGB edge density, a proxy for 2D clutter. Both respond to object boundaries, but their Spearman correlation is modest ($\rho = 0.298$). Grouping samples into thirds by each measure gives 41.3\% agreement (33.3\% expected by chance). Depth layer count, which defines our complexity strata, is uncorrelated with RGB edge density ($\rho = 0.042$), suggesting that these strata do not simply track increasing visual clutter.

Shared-anchor density (\Cref{fig:anchor_density}) is not significantly correlated with depth layer count ($\rho = 0.115$, $p = 0.23$) and is only weakly correlated with depth edge density ($\rho = 0.242$, $p = 0.012$). Together, these checks suggest that the depth analysis is not simply measuring RGB-based visual clutter and that the depth and shared-landmark analyses capture different aspects of scene complexity.

\subsection{Explicit Depth Input and a Specialized Spatial Model}
\label{sec:supp_depth_input}

We complement the RGB-only analysis by testing explicit depth input in general-purpose MLLMs and a specialized spatial model.

\noindent \textbf{Depth Maps as Input.} We append metric depth maps from Depth-Anything-V2-Metric-Indoor-Large~\cite{yang2024depth} after the RGB views of the 472 two-agent Distance Awareness questions, keeping the prompt unchanged. Adding depth provides no clear benefit across the tested models (\Cref{tab:supp_depth_input}). Gemini-2.5-Pro and Qwen2.5-VL-7B drop by about 13 percentage points, while the two LLaVA models change little. Replacing the depth maps with duplicated RGB views reduces Qwen2.5-VL-7B accuracy by only 1.8 points. This suggests that its larger drop with depth cannot be explained by the number of additional images alone.

For Qwen2.5-VL-7B, the drop from adding depth is comparable to the 17.2-point drop from removing one agent's view in our observation-removal ablation on the same questions. In contrast, LLaVA-NeXT-Video-7B changes little under either condition ($+0.6$ and $+0.4$ points, respectively). In neither case does depth information provide a clear improvement, supporting our finding that additional per-view depth cues do not address the core bottleneck in multi-agent reasoning.

\begin{table}[t]
\centering
\caption{Distance Awareness accuracy (\%) without and with metric depth maps. Scores use questions with valid responses in both conditions: 470 for Gemini-2.5-Pro, 467 for LLaVA-OneVision-7B, and 472 for the other models. $\Delta$ is the change in percentage points, computed before rounding; $p$ is from a paired McNemar test.}
\label{tab:supp_depth_input}
\setlength{\tabcolsep}{6pt}
\renewcommand{\arraystretch}{0.95}
{\small
\begin{tabular}{lcccc}
    \toprule
    \textbf{Model} & \textbf{RGB only} & \textbf{+Depth} & \textbf{\boldmath$\Delta$} & \textbf{\boldmath$p$} \\
    \midrule
    Gemini-2.5-Pro      & 61.9 & 48.3 & $-13.6$ & $<10^{-5}$ \\
    Qwen2.5-VL-7B       & 53.2 & 39.8 & $-13.3$ & $<10^{-5}$ \\
    LLaVA-OneVision-7B  & 49.7 & 47.1 & $-2.6$  & 0.25 \\
    LLaVA-NeXT-Video-7B & 46.2 & 46.8 & $+0.6$  & 0.82 \\
    \bottomrule
\end{tabular}}
\end{table}

\noindent \textbf{Specialized Spatial Model.} We evaluate SpatialBot~\cite{cai2025spatialbot}, a model designed for spatial reasoning with RGB and depth input, on the two-agent subset (2,590 questions). Using its original implementation, we provide both agent views with depth maps from ZoeDepth~\cite{bhat2023zoedepthzeroshottransfercombining}. We compare it with LLaVA-OneVision-7B and Gemini-2.5-Flash, which receive RGB views only (\Cref{tab:supp_spatialbot}). Despite its spatial specialization and access to explicit depth, SpatialBot performs below both general-purpose models in every task and overall (33.2\% vs.\ 40.5\% and 44.1\%). These results suggest that spatial specialization and access to depth do not, on their own, ensure strong performance in MaRiO's multi-agent setting. They also reinforce our central finding that the primary bottleneck is reasoning about cross-agent viewpoint correspondence.

\begin{table}[t]
\centering
\caption{Accuracy on the two-agent subset (2,590 questions). SpatialBot receives RGB views and depth maps; the other models receive RGB views only.}
\label{tab:supp_spatialbot}
\setlength{\tabcolsep}{6pt}
\renewcommand{\arraystretch}{0.95}
{\small
\begin{tabular}{llcccccc}
    \toprule
    \textbf{Model} & \textbf{Input} & \textbf{Ego.} & \textbf{Nav.} & \textbf{Dist.} & \textbf{Occ.} & \textbf{Rel.} & \textbf{Overall} \\
    \midrule
    LLaVA-OneVision-7B & RGB         & 30.4 & 36.4 & 49.8 & 47.8 & 43.2 & 40.5 \\
    Gemini-2.5-Flash   & RGB         & 31.3 & 44.6 & 51.5 & 50.3 & 46.3 & 44.1 \\
    SpatialBot        & RGB + depth & 27.1 & 24.6 & 37.1 & 42.0 & 41.3 & 33.2 \\
    \bottomrule
\end{tabular}}
\end{table}

\subsection{Impact of Geometry-Guided Synthetic View Augmentation}
\Cref{tab:supp_app} shows adding synthetic intermediate views yields consistent improvements across most models, with particularly large gains for models that initially exhibited weaker performance. This suggests that models struggling with multi-view spatial reasoning may benefit more from additional intermediate observations that help bridge viewpoint gaps and provide smoother visual transitions between agents. 

\subsection{Ablation of Synthetic Views}
\label{sec:supp_ablation}

To complement the main results in \Cref{tab:main_app}, we ablate three design choices in the synthetic view augmentation pipeline: the number of frames sampled from the generated video, the order in which agent viewpoints are interpolated, and whether gains arise from novel views or simply from additional frames. All ablations use Qwen2.5-VL-7B-Instruct as the base model.

\begin{figure}[t!]
    \centering
    \includegraphics[width=.46\linewidth]{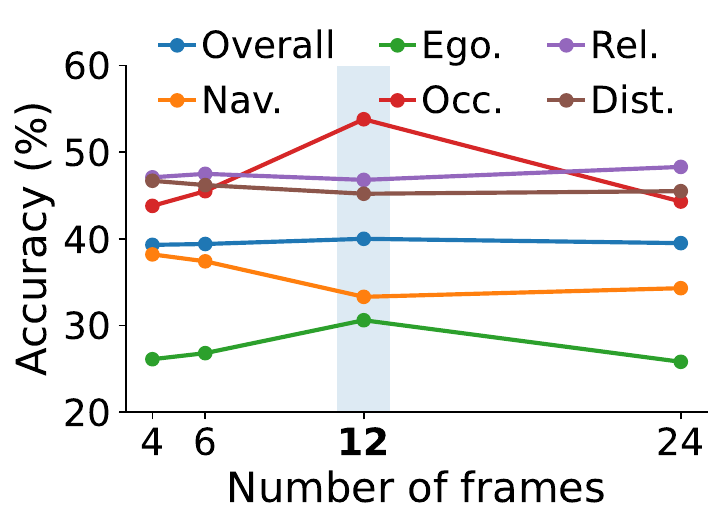}
    \hfill
    \includegraphics[width=.46\linewidth]{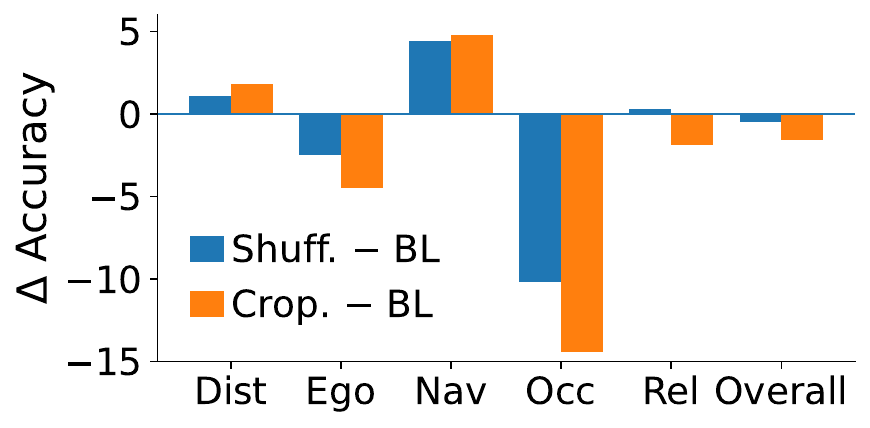}
    \caption{\textbf{Ablation of generated synthetic views.} \textit{Left:} effect of the number of synthetic frames sampled from the generated video. \textit{Right:} effect of interpolation order (Shuff. $-$ BL) and image augmentation with cropped frames (Crop. $-$ BL), both reported as accuracy deltas relative to the default baseline.}
    \label{fig:ana_app}
\end{figure}

\noindent \textbf{Number of Frames.} Performance varies with the number of frames sampled from the generated synthetic video. As shown in \Cref{fig:ana_app} (\textit{left}), the model achieves the best overall performance when using 12 frames. Interestingly, the distance awareness score decreases as the number of frames increases, possibly because additional frames introduce visual redundancy that confuses the model when estimating relative distances from multiple viewpoints. Results for the remaining categories remain inconclusive.

\noindent \textbf{Order of Interpolation.} We observe minimal variation in performance when changing the order of interpolated frames. In our default setting, videos are generated by interpolating viewpoints in the order \textit{agent 1} $\rightarrow$ \textit{agent 2} $\rightarrow$ \textit{agent 3}. For the shuffled setting, we randomly permute the agent order before generating the interpolated video. The resulting scores remain largely unchanged (\Cref{fig:ana_app}, \textit{right}), suggesting that the interpolation order has little impact on model performance.

\noindent \textbf{Image Augmentation.} To determine whether improvements arise from the novel viewpoints or simply from providing more frames, we replace interpolated views with cropped frames from the original images. This augmentation does not outperform the baseline configuration (\Cref{fig:ana_app}, \textit{right}), suggesting that the gains primarily come from novel synthesized viewpoints rather than from additional frames alone. A small improvement is observed only for navigation.

\begin{table}[t!]
    \centering
    \caption{\textbf{Performance of synthetic view augmentations.} \colorbox{improved}{Improvements} after adding synthetic views and \textbf{\textcolor{darkgreen}{best}} overall gain are highlighted. $\dag$ denotes the corrected average performance; the corresponding value in \Cref{tab:main_results} contained a minor typographical error.}
    \begin{tabular}{l|cc|ccc|c|c}
        \toprule
        \multirow{2}{*}{\textbf{Model}} & \multicolumn{2}{c|}{\textbf{Comp.}} & \multicolumn{3}{c|}{\textbf{Static Perception}} & \multirow{2}{*}{\textbf{Overall}} & \multirow{2}{*}{\textbf{Avg.}}\\
        & \textbf{Ego.} & \textbf{Nav.} & \textbf{Dist.} & \textbf{Occ.} & \textbf{Rel.} & & \\
        \midrule
        \rowcolor{blue!5}\multicolumn{8}{l}{\textbf{General Models}} \\
        Phi-3.5-Vision-Instruct-4B & 27.3 & 31.9 & 36.7 & 37.3 & 31.4 & 32.3 & 32.9 \\
        \hspace{2em}$+$ Syn. View & \cellcolor{improved}28.4 & \cellcolor{improved}34.4 & \cellcolor{improved}38.7 & \cellcolor{improved}39.2 & \cellcolor{improved}34.1 & \cellcolor{improved}34.3 & \cellcolor{improved}35.0\\
        & \textcolor{darkgreen}{\scriptsize{+1.1}} & \textcolor{darkgreen}{\scriptsize{+2.5}} & \textcolor{darkgreen}{\scriptsize{+2.0}} & \textcolor{darkgreen}{\scriptsize{+1.9}} & \textcolor{darkgreen}{\scriptsize{+2.7}} & \textcolor{darkgreen}{\scriptsize{+2.0}} & \textcolor{darkgreen}{\scriptsize{+2.1}}\\
        \midrule
        VILA1.5-7B & 28.1 & 32.5 & 41.3 & 39.8 & 37.5 & 34.9 & 35.8 \\
        \hspace{2em}$+$ Syn. View & 27.5 & \cellcolor{improved}36.7 & 41.1 & \cellcolor{improved}44.2 & \cellcolor{improved}43.8 & \cellcolor{improved}37.6 & \cellcolor{improved}38.6\\
        & \textcolor{red}{\scriptsize-0.6} & \textcolor{darkgreen}{\scriptsize{+4.2}} & \textcolor{red}{\scriptsize-0.2} & \textcolor{darkgreen}{\scriptsize{+4.4}} & \textcolor{darkgreen}{\scriptsize{+6.3}} & \textcolor{darkgreen}{\scriptsize{+2.7}} & \textcolor{darkgreen}{\scriptsize{+2.8}}\\
        \midrule
        InternVL3-2B & 29.2 & 32.9 & 37.9 & 39.7 & 33.6 & 34.0 & 34.6 $\dag$ \\
        \hspace{2em}$+$ Syn. View & \cellcolor{improved}30.4 & 31.2 & \cellcolor{improved}45.5 & 38.0 & \cellcolor{improved}35.6 & \cellcolor{improved}35.4 & \cellcolor{improved}36.1\\
        & \textcolor{darkgreen}{\scriptsize{+1.2}} & \textcolor{red}{\scriptsize-1.7} & \textcolor{darkgreen}{\scriptsize{+7.6}} & \textcolor{red}{\scriptsize-1.7} & \textcolor{darkgreen}{\scriptsize{+2.0}} & \textcolor{darkgreen}{\scriptsize{+1.4}} & \textcolor{darkgreen}{\scriptsize{+1.5}}\\
        \bottomrule
    \end{tabular}
    \label{tab:supp_app}
\end{table}

\begin{figure}[t!]
  \centering
  \includegraphics[width=0.7\linewidth]{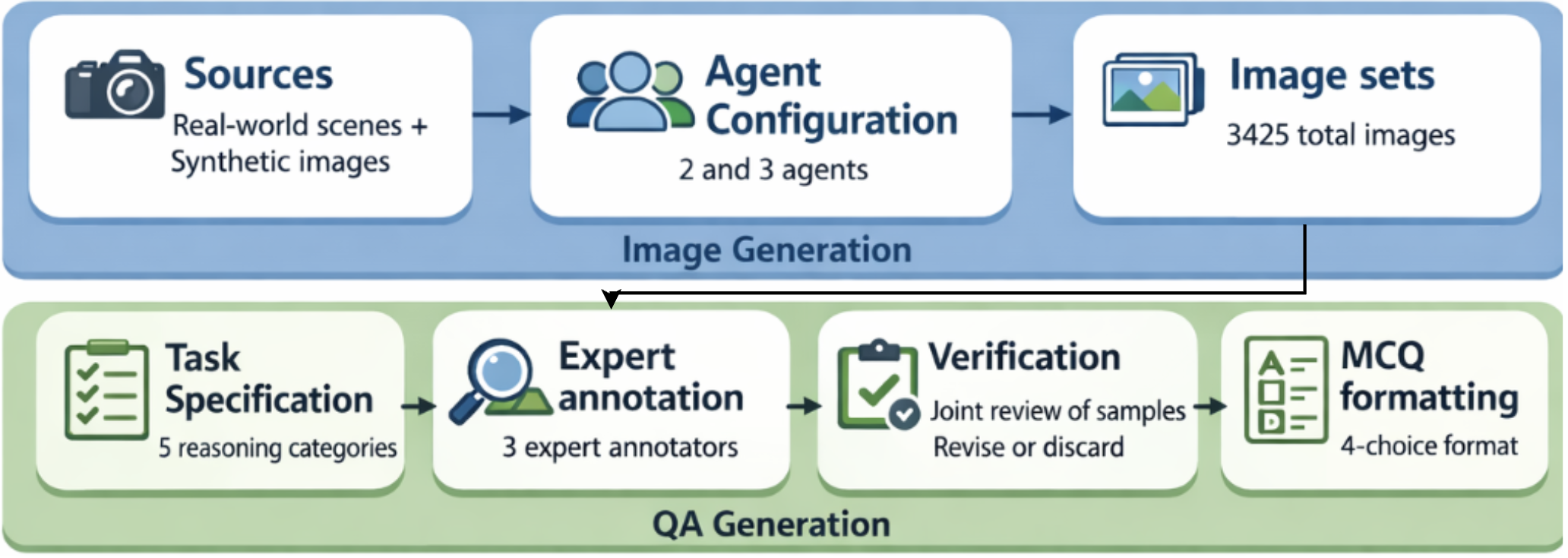}
  \caption{\textbf{MaRiO construction pipeline.}
  Image Generation (top) combines real-world and synthetic sources into multi-agent observation sets via agent configuration. QA Generation (bottom) produces verified 4-choice questions through task specification, expert annotation, joint verification, and MCQ formatting.}
  \label{fig:construction_pipeline}
\end{figure}
\begin{figure}[t!]
    \centering
    \includegraphics[width=\linewidth]{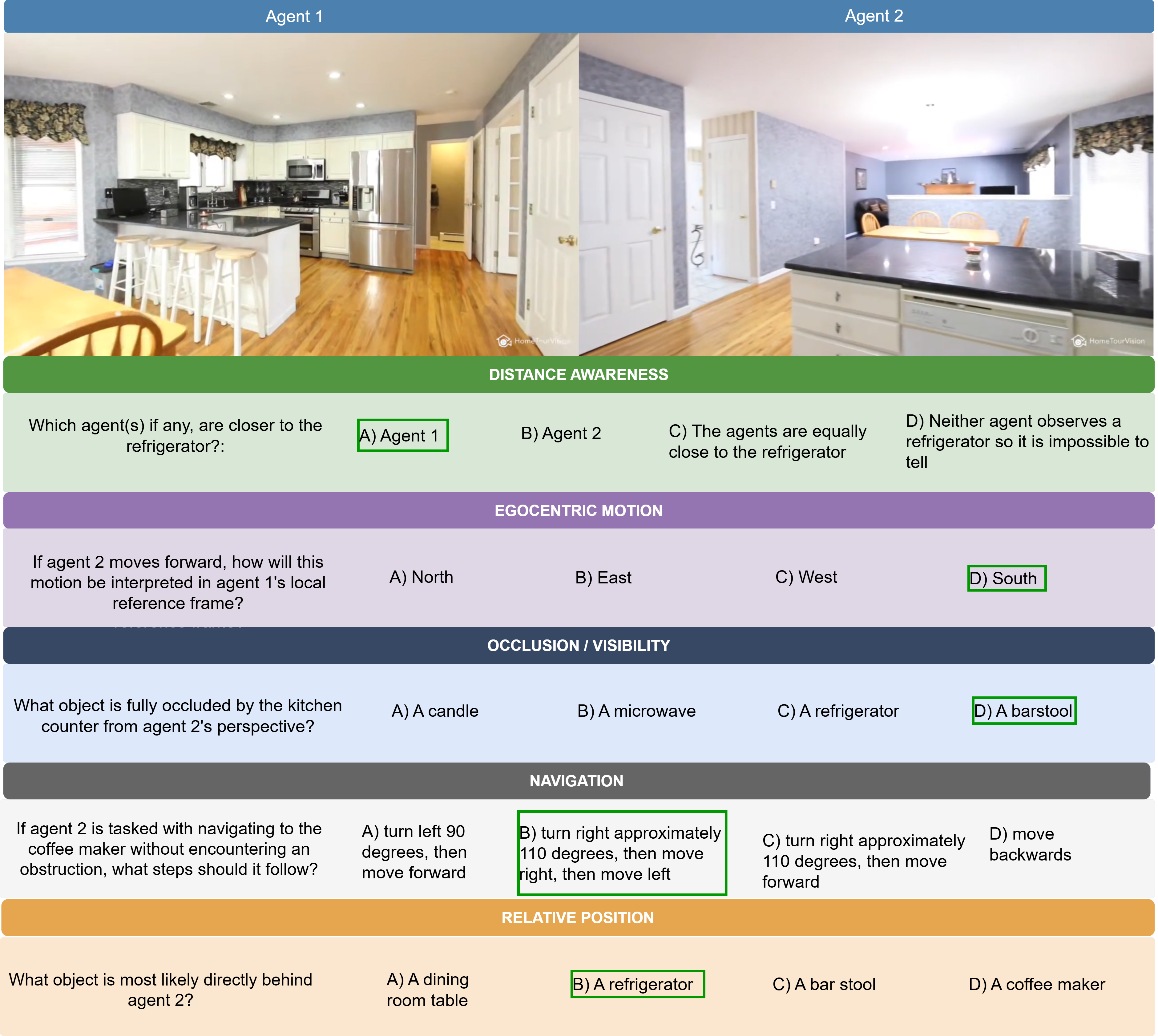}
    \caption{\textbf{Multi-agent observations covering all categories of reasoning tasks of MaRiO.}}
    \label{fig:multi_cat_example}
\end{figure}

\section{Benchmark Construction and Validation}
\label{sec:bench-supp}
\subsection{Generation Pipeline}
Figure~\ref{fig:construction_pipeline} illustrates the MaRiO construction pipeline, which proceeds in two stages: Image Generation and QA Generation. 
During the agent configuration stage for image generation, each scene is assigned either a two-agent or three-agent setup by selecting independent egocentric viewpoints observing the same environment. Viewpoints are chosen to ensure spatial overlap while maintaining distinct perspectives, such that solving the task requires integrating information across multiple views rather than relying on a single observation. For question generation, the task specifications are defined through the system prompts described in the next section. Table~\ref{tab:category_distribution} summarizes the resulting distribution of samples across categories and agent configurations. 
\Cref{fig:multi_cat_example} shows an example of multi-agent collaborative observation covering reasoning tasks across all categories.

\begin{figure}[t!]
    \centering
    \includegraphics[width=\linewidth, height=18cm]{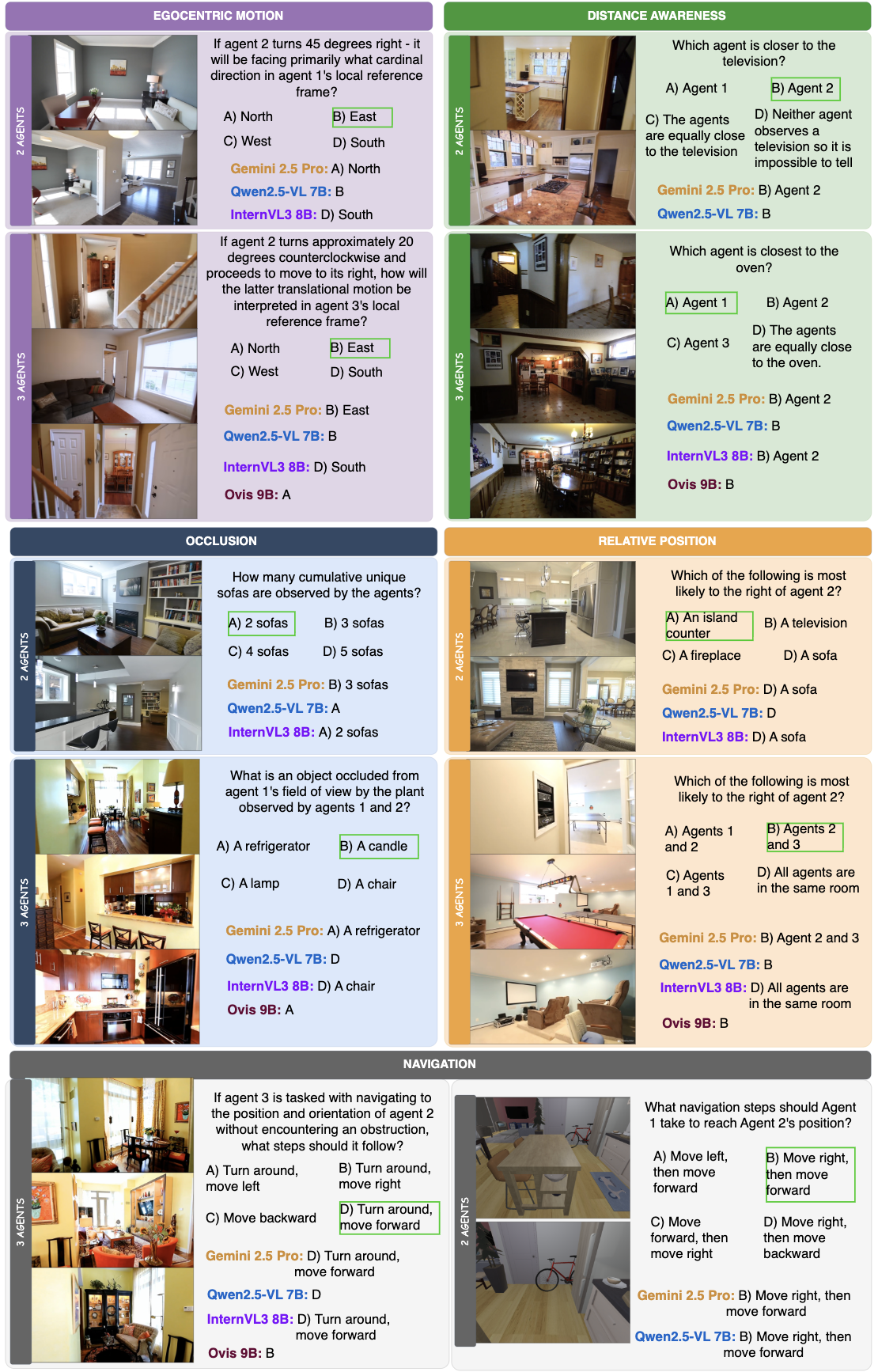}
    \caption{\textbf{Qualitative analysis of MaRiO.} Here, ground truth option is highlighted in green box, followed by predictions of several models.}
    \label{fig:supp_samples}
\end{figure}

\clearpage
\begin{table}[t!]
  \centering
  \caption{\textbf{Category-wise distribution of MaRiO.}
  Comp.\ and Static Perc.\ respectively denote Compositions of Geometric Transformations and Static Spatial Perception.}
  \label{tab:category_distribution}
  \setlength{\tabcolsep}{5pt}
      \begin{tabular}{@{}llrrr@{}}
        \toprule
        \textbf{Type} & \textbf{Category} & \textbf{2-Agents} & \textbf{3-Agents} & \textbf{Total} \\
        \midrule
        \multirow{2}{*}{Comp.}
          & Egocentric Motion  & 576 & 189 & 765 \\
          & Navigation         & 659 & 214 & 873 \\
        \midrule
        \multirow{3}{*}{Static Perc.}
          & Distance Awareness & 472 & 149 & 621 \\
          & Occlusion          & 366 &  41 & 407 \\
          & Relative Position  & 517 &  72 & 589 \\
        \midrule
          & \textbf{All}       & \textbf{2,590} & \textbf{665} & \textbf{3,255} \\
        \bottomrule
    \end{tabular}
\end{table}

\subsection{System Prompts}
Here we present the prompts used for each task category in the benchmark. All evaluations follow a shared master prompt that enforces consistent answer formatting and instructs the model to base its decision strictly on the visual evidence across agents’ observations. In addition, each reasoning category is paired with a specialized prompt that defines the spatial semantics required for that task, including assumptions about camera placement, agent motion, egocentric coordinate systems, occlusion definitions, and relative agent positioning. These category-specific prompts ensure that models interpret spatial relations, navigation primitives, and visibility conditions under a consistent and well-defined set of rules during evaluation.

\begin{promptbox}{Master Prompt}
    You are an AI assistant performing an academic benchmark evaluation.
    The question has exactly four answer options presented in alphabetical order (A, B, C, D).
    You MUST respond with exactly ONE alphabet character in the set {A, B, C, D}. Select the character that corresponds to the most plausible answer.
    Do NOT provide explanations, reasoning, analysis, or any additional text.
    Do NOT include multiple answers.
    Your response must consist of a single line only.
    The answer MUST be based strictly on the visual evidence in the images.
    If more than one option appears plausible, select the single most consistent option with the given views. \newline \newline

    There are $N$ agents operating within the same scene. The first image was captured by agent 1, and the second image was captured by agent 2, ..., and the $N^{th}$ image was captured by agent $N$. The images were captured by a camera with exactly the same intrinsic settings, and the images were captured at exactly the same time.
\end{promptbox}

\begin{promptbox}{Distance Awareness}
    Both agents are at the same height unless mentioned otherwise. The only difference between the two images is the agent's 3D position.
\end{promptbox}

\begin{promptbox}{Egocentric/Allocentric Motion}
    In this scene, cardinal directions will be bound to the local coordinate systems defined by agents' egocentric frames of reference. The camera is positioned on and oriented along the agent's forward axis. The agent's forward-facing direction is North. The opposite direction (extending outward from the back of the agent) is South. Extending outward from the agent's right side is East. Extending outward from the agent's left side is west.
\end{promptbox}
\newpage

\begin{promptbox}{Navigation}
     Assume that the camera is fixed to the front of the agent and faces outwards along the agent's forward axis. Thus, changes to the agent's position and orientation by way of locomotion are equivalently applied to the camera's position and orientation. If the agent is tasked with turning left or turning right, assume that these commands refer to rotating in place 90 degrees towards left and rotating in place 90 degrees towards right respectively. Assume that turning around refers to rotating in place 180 degrees. Assume that prompts such as 'turn slightly left' and 'turn slightly right' refer to rotating in place less than 90 degrees to the left and rotating in place less than 90 degrees to the right respectively.    
\end{promptbox}

\begin{promptbox}{Occlusion/Visibility/Counting}
    'Fully occluded' means the object cannot be seen at all. 'Partially occluded' means the object can be partially hidden, but still detectable. Generally speaking, the use of 'occluded' can be interpreted as either 'fully occluded' OR 'partially occluded'.  
\end{promptbox}

\begin{promptbox}{Relative Agents}
    Position Semantics are defined as follows: 'Ahead of' = Along the same direction as the agent's camera is facing; 'Behind' = Along the opposite direction as the agent's camera is facing; 'Left-side' = Perpendicular to the agent's viewing direction on the agent's left side; 'Right-side' = Perpendicular to the agent's viewing direction on the agent's right side.
\end{promptbox}

\paragraph{Agent--Image Association.} The association between images and agents is communicated to the model through the text prompt. The ``Agent $x$'' labels shown in the figures are illustrative and are not embedded in the images. As specified in the master prompt, the $i^{th}$ image is stated to have been captured by agent $i$, so the mapping is always given explicitly rather than inferred from image order alone. For interfaces that support interleaved image--text inputs, we reinforce this mapping with adjacent textual references (e.g., ``Image 1: \texttt{<image>}''). For interfaces that do not, images are passed in order with the same prompt-based mapping, as the preprocessor preserves image order.

For the synthetic view augmentation experiments, the generated intermediate views are not associated with any agent. They are provided as auxiliary visual context showing the relationship between agents' viewpoints, and the prompt states explicitly that they are supporting context rather than an additional agent observation.

\paragraph{Agent Visibility.} Agents are not directly visible to one another, reflecting scenarios in which they operate beyond visual line of sight. The model must therefore infer their relative positions and orientations from correspondences across their views.

\subsection{Annotation Protocol}
\label{sec:annotation_protocol}

Question generation was done by three expert annotators drawn from the author team. Each annotator was assigned primary responsibility for one or more image sources (RealEstate10K, NAVI, VirtualHome, Habitat-Sim with Replica CAD, or AI2-THOR) and also contributed to other sources. Task types were not partitioned across annotators; instead, all annotators contributed questions across the five reasoning categories.

The annotation pipeline proceeded in three stages. Annotators first drafted candidate questions following the per-category specifications, ensuring that each question required integrating evidence across multiple agents' views. Multiple-choice options were then drafted with LLM assistance, it is an LLM was prompted to propose three plausible distractors per question, which an annotator reviewed, edited, or replaced. The ground-truth answer for every sample was authored by a human; no LLM output was accepted as a correct answer.

Each draft question was reviewed by at least one additional annotator, and occasionally by two, for spatial validity, ambiguity, and consistency with the category-level specification. When a reviewer flagged a question as ambiguous, the team first tried to repair it by reformulating the prompt over the same observation set; the small minority of unrecoverable cases were discarded. Disagreements during review were resolved through consensus discussion among the three annotators.

To examine the consistency of QA style across annotators, we conduct a blind audit on a sample of the benchmark. We sample 25 QA pairs per annotator per category (375 items in total) and have an LLM judge (GPT-5.2) rate each item against four binary criteria, with annotator identity hidden from the judge: (i) whether the question asks the kind of thing its reasoning category is meant to ask; (ii) whether its four answer options share a consistent format; (iii) whether it unambiguously names the scene entities it depends on; and (iv) whether its phrasing, framing, and level of detail match the overall style of the dataset.

Table~\ref{tab:annotation_audit} reports the pass rate for each criterion. All three annotators pass at high rates across all criteria (92.8--98.8\% overall, 94.9\% pooled). Pass rates are highest for answer-option consistency (97.3\%) and entity concreteness (97.1\%), and lowest for overall style match (91.2\%), where the gap across annotators is also largest (86.4--97.6\%).

\begin{table}[t]
\centering
\caption{Blind audit of QA-style consistency across the three annotators (A1--A3), reported as the percentage of sampled items passing each criterion (125 items per annotator).}
\label{tab:annotation_audit}
\setlength{\tabcolsep}{6pt}
\renewcommand{\arraystretch}{0.95}
{\small
\begin{tabular}{lcccc}
    \toprule
    \textbf{Criterion} & \textbf{A1} & \textbf{A2} & \textbf{A3} & \textbf{Pooled} \\
    \midrule
    (i) Category-style fit          & 99.2  & 92.8 & 89.6 & 93.9 \\
    (ii) Answer-option consistency  & 100.0 & 92.8 & 99.2 & 97.3 \\
    (iii) Entity concreteness       & 98.4  & 96.8 & 96.0 & 97.1 \\
    (iv) Overall style match        & 97.6  & 89.6 & 86.4 & 91.2 \\
    \midrule
    \textbf{Overall}                & \textbf{98.8} & \textbf{93.0} & \textbf{92.8} & \textbf{94.9} \\
    \bottomrule
\end{tabular}}
\end{table}

\subsection{Human Evaluation Protocol}
\label{sec:human_eval}
The human-baseline accuracy reported in Table~\ref{tab:main_results} was collected on a stratified subset of MaRiO. We sampled 97 questions in total: 17 from Distance Awareness, 21 from Egocentric Motion, 16 from Navigation, 22 from Occlusion/Visibility, and 21 from Relative Position. Within each category, the ratio of 2-agent to 3-agent questions was kept proportional to the full benchmark.

Responses were collected anonymously from students in computer science, engineering, or related fields, none of whom were members of the author team. Each evaluator received the same master prompt presented to the MLLMs, with identical images, answer-option formatting, and the instruction to return a single answer letter. No time limit was set, and no auxiliary information beyond the RGB inputs was given.

\subsection{Distractor Construction Rules}
\label{sec:distractor_rules}

To maintain consistent distractor quality across reasoning categories, we followed a category-specific construction scheme.

\paragraph{Distance Awareness.} The answer space includes the agent identifiers, along with edge-case options such as all agents are equally close or none of the agents observe the target. Distractors are selected from the remaining valid agents or edge-case options.

\paragraph{Egocentric Motion.} Options are generally restricted to the four cardinal directions, e.g., North, East, South, and West. Distractors are selected from the remaining directions that do not correspond to the ground-truth direction.

\paragraph{Navigation.} The answer space consists of motion sequences over a fixed primitive set $\{$move forward, move backward, turn left, turn right, turn around$\}$. Distractors are constructed by applying minimal directional perturbations to the ground-truth sequence, that is done by inverting one or more direction primitives or reordering steps, so that each distractor remains a syntactically valid navigation plan but fails to satisfy the goal.

\paragraph{Occlusion / Visibility.} For object-identification questions, the answer space consists of objects present in the scene; distractors are drawn from objects visible in some agent's view but not satisfying the queried visibility condition. For counting questions, distractors are integers within $\pm 1$ or $\pm 2$ of the ground-truth count.

\paragraph{Relative Position.} The answer space consists of objects appearing in the combined visual context of the agent set. Distractors are drawn from objects positioned in directions adjacent to, or otherwise confusable with, the queried direction relative to the reference agent.

\subsection{Number of Answer Options}
\label{sec:supp_choices}

To analyze how model performance on MaRiO scales with the size of the answer space, we use GPT-5.2 to expand 600 existing QA samples from four to six and eight answer options, keeping the original questions and correct answers unchanged. The results show a consistent decline in overall accuracy as the number of answer options increases across all three evaluated models (\Cref{tab:supp_choices_gemini,tab:supp_choices_llava,tab:supp_choices_qwen}). The magnitude of this decline varies across tasks. Distance Awareness has fewer samples in this analysis because its more constrained answer space limits how far it can be expanded. These results highlight sensitivity to the number of distractors, a limitation of the multiple-choice evaluation format.

\begin{table}[t]
\centering
\caption{Gemini-2.5-Flash accuracy across answer-choice settings.}
\label{tab:supp_choices_gemini}
\setlength{\tabcolsep}{6pt}
\renewcommand{\arraystretch}{0.95}
{\small
\begin{tabular}{lccc}
    \toprule
    \textbf{Task} & \textbf{4 Choices} & \textbf{6 Choices} & \textbf{8 Choices} \\
    \midrule
    Distance Awareness & 35.90\% (14/39)   & 38.46\% (15/39)   & 33.33\% (13/39) \\
    Egocentric Motion  & 40.57\% (43/106)  & 23.58\% (25/106)  & 20.75\% (22/106) \\
    Navigation         & 39.11\% (79/202)  & 31.68\% (64/202)  & 28.22\% (57/202) \\
    Occlusion          & 44.44\% (48/108)  & 32.41\% (35/108)  & 25.00\% (27/108) \\
    Relative Position  & 44.14\% (64/145)  & 30.34\% (44/145)  & 19.31\% (28/145) \\
    \midrule
    Overall            & 41.33\% (248/600) & 30.50\% (183/600) & 24.50\% (147/600) \\
    \bottomrule
\end{tabular}}
\end{table}

\begin{table}[t]
\centering
\caption{LLaVA-OneVision-7B accuracy across answer-choice settings.}
\label{tab:supp_choices_llava}
\setlength{\tabcolsep}{6pt}
\renewcommand{\arraystretch}{0.95}
{\small
\begin{tabular}{lccc}
    \toprule
    \textbf{Task} & \textbf{4 Choices} & \textbf{6 Choices} & \textbf{8 Choices} \\
    \midrule
    Distance Awareness & 20.51\% (8/39)    & 28.21\% (11/39)   & 17.95\% (7/39) \\
    Egocentric Motion  & 30.19\% (32/106)  & 16.98\% (18/106)  & 20.75\% (22/106) \\
    Navigation         & 36.14\% (73/202)  & 29.70\% (60/202)  & 19.80\% (40/202) \\
    Occlusion          & 51.85\% (56/108)  & 34.26\% (37/108)  & 26.85\% (29/108) \\
    Relative Position  & 50.34\% (73/145)  & 24.83\% (36/145)  & 20.69\% (30/145) \\
    \midrule
    Overall            & 40.33\% (242/600) & 27.00\% (162/600) & 21.33\% (128/600) \\
    \bottomrule
\end{tabular}}
\end{table}

\begin{table}[t]
\centering
\caption{Qwen2.5-VL-7B accuracy across answer-choice settings.}
\label{tab:supp_choices_qwen}
\setlength{\tabcolsep}{6pt}
\renewcommand{\arraystretch}{0.95}
{\small
\begin{tabular}{lccc}
    \toprule
    \textbf{Task} & \textbf{4 Choices} & \textbf{6 Choices} & \textbf{8 Choices} \\
    \midrule
    Distance Awareness & 35.90\% (14/39)   & 41.03\% (16/39)   & 41.03\% (16/39) \\
    Egocentric Motion  & 33.96\% (36/106)  & 26.42\% (28/106)  & 21.70\% (23/106) \\
    Navigation         & 33.17\% (67/202)  & 26.24\% (53/202)  & 25.74\% (52/202) \\
    Occlusion          & 50.93\% (55/108)  & 31.48\% (34/108)  & 39.81\% (43/108) \\
    Relative Position  & 45.52\% (66/145)  & 19.31\% (28/145)  & 15.17\% (22/145) \\
    \midrule
    Overall            & 39.67\% (238/600) & 26.50\% (159/600) & 26.00\% (156/600) \\
    \bottomrule
\end{tabular}}
\end{table}

\subsection{Task Taxonomy}
\label{sec:task_taxonomy}
To understand how the five tasks of MaRiO relate to one another, we computed pairwise Spearman rank correlations of task-wise accuracy across the 19 open-source models on the full benchmark (Table~\ref{tab:task_corr}). Distance Awareness, Occlusion, and Relative Position are strongly correlated with one another (0.855--0.933), whereas Egocentric Motion and Navigation correlate more weakly with these tasks (0.399--0.634) and with each other (0.407).

This pattern led us to group the tasks in relevant analyses as \textit{Static Perception} (Distance Awareness, Occlusion, and Relative Position) and \textit{Compositional} (Egocentric Motion and Navigation). The grouping reflects the type of reasoning each task requires rather than its content. Static Perception tasks are answered from perceptual cues identified across the agents' views, such as relative proximity, visibility, and relative direction. Since MaRiO provides only RGB observations without camera poses or coordinates, this includes Relative Position, which must be inferred from perceptual cues rather than computed from known viewpoint relationships. Compositional tasks additionally require composing sequential viewpoint transformations or motions across reference frames.

\begin{table}[t]
\centering
\caption{Spearman rank correlation of model accuracy across task categories, computed over the 19 open-source models on the full benchmark.}
\label{tab:task_corr}
\setlength{\tabcolsep}{6pt}
\renewcommand{\arraystretch}{0.95}
{\small
\begin{tabular}{l|cc|ccc}
    \toprule
     & \multicolumn{2}{c|}{\textbf{Compositional}} & \multicolumn{3}{c}{\textbf{Static Perception}} \\
     & \textbf{Ego.} & \textbf{Nav.} & \textbf{Dist.} & \textbf{Occ.} & \textbf{Rel.} \\
    \midrule
    Ego. & 1.000 & 0.407 & 0.399 & 0.530 & 0.634 \\
    Nav. & 0.407 & 1.000 & 0.426 & 0.612 & 0.542 \\
    \midrule
    Dist. & 0.399 & 0.426 & 1.000 & 0.933 & 0.855 \\
    Occ.  & 0.530 & 0.612 & 0.933 & 1.000 & 0.856 \\
    Rel.  & 0.634 & 0.542 & 0.855 & 0.856 & 1.000 \\
    \bottomrule
\end{tabular}}
\end{table}

\begin{table}[t]
\centering
\caption{Accuracy (\%) on synthetic and real-world image sets.}
\label{tab:syn_vs_real}
\begin{tabular}{lcc}
\toprule
Model & Synthetic & Real-World \\
\midrule
LLaVA-NeXT-Video-7B-HF   & 31.97 & 32.61 \\
mPLUG-Owl3-7B-240728     & 34.77 & 37.90 \\
Qwen2-VL-7B-Instruct     & 39.86 & 42.97 \\
Qwen2.5-VL-3B-Instruct   & 40.76 & 40.81 \\
Qwen2.5-VL-7B-Instruct   & 37.26 & 39.71 \\
\bottomrule
\end{tabular}
\end{table}
\subsection{Real vs. Synthetic Data}
\label{sec:real_vs_syn}
To examine how image source affects performance, we evaluate each model separately on real-world and synthetic image sets. The real-world subset contains 913 image sets drawn from RealEstate10K and NAVI, and the synthetic subset contains 767 image sets generated with VirtualHome, Habitat-Sim, and AI2-THOR. Table~\ref{tab:syn_vs_real} reports accuracy for each subset.

Most models score comparably or slightly higher on real-world image sets than on synthetic ones. The gap is small for most models, with mPLUG-Owl3-7B showing the largest difference ($\Delta 3.13\%$). Qwen2.5-VL-3B-Instruct performs nearly identically across the two sources. Overall, relative model performance is consistent across data sources, and neither source is uniformly easier than the other.

\section{Qualitative Analysis}
\label{sec:qual-supp}
\Cref{fig:supp_samples} presents representative examples across the five reasoning categories, highlighting common success and failure patterns among models. Closed-source models generally demonstrate stronger performance in tasks requiring coordinated spatial reasoning across multiple views, particularly in distance awareness and navigation where integrating observations from different agents is critical. In contrast, many open-source models exhibit inconsistent predictions, often relying on partial visual cues from a single viewpoint rather than reconciling information across agents. This behavior becomes more pronounced in the 3-agent setting, where the additional viewpoint increases the need for coherent cross-view alignment. For instance, tasks involving occlusion reasoning and relative positioning frequently require identifying objects that are only partially visible across views; while some models correctly aggregate these cues, others fail to maintain consistent correspondences between observations. Conversely, egocentric motion questions show mixed behavior across both model types, suggesting that directional reasoning remains challenging even when visual ambiguity is limited. Overall, these examples illustrate that while some models can leverage multi-agent observations effectively, many still struggle to consolidate distributed visual evidence into a unified spatial representation, particularly as the number of agents increases.

\section{Future Research Direction}
\label{sec:future}
MaRiO focuses on synchronized multi-agent observations captured at the same time and under consistent environmental conditions for multi-agent collaborative reasoning. An important direction for future work is to explore time-asynchronous observations, where agents may observe the same environment at different times. For example, one agent may capture the scene during the day while another observes it at night, or agents may view the same location across different seasons such as summer and winter. Such scenarios introduce appearance shifts while preserving the underlying spatial structure, creating additional challenges for models to maintain consistent spatial reasoning despite changes in lighting, weather, or scene dynamics. Another promising direction is developing models that explicitly maintain structured representations across independently oriented viewpoints. Our findings primarily suggest that current MLLMs struggle to integrate distributed observations into a coherent scene-level understanding. While we attempted to partially address this by augmenting geometry-guided intermediate transition views between agents’ independently oriented observations, the resulting improvements were modest for comparatively stronger models. This suggests that simply augmenting observations may not fully resolve the underlying limitation. Instead, future work may require architectures that explicitly model cross-view correspondences or maintain scene-level spatial memory in order to effectively consolidate multi-agent observations.

\newpage
\section*{NeurIPS Paper Checklist}


\begin{enumerate}

\item {\bf Claims}
    \item[] Question: Do the main claims made in the abstract and introduction accurately reflect the paper's contributions and scope?
    \item[] Answer: \answerYes{}
    \item[] Justification: The abstract and Section~1 (Introduction) outline our three contributions: the MaRiO benchmark for multi-agent collaborative reasoning, a zero-shot evaluation of 23 state-of-the-art MLLMs, and a geometry-guided synthetic view augmentation. These claims correspond directly to the experimental scope reported in Sections~3-5 and the findings summarized in Section~6.

\item {\bf Limitations}
    \item[] Question: Does the paper discuss the limitations of the work performed by the authors?
    \item[] Answer: \answerYes{}
    \item[] Justification: Limitations are discussed in Section~6 (Conclusion, ``Limitations'' paragraph). We acknowledge that MaRiO targets static, snapshot-based observations in a multiple-choice QA format and does not capture temporal dynamics, asynchronous observations, or interactive decision-making.

\item {\bf Theory assumptions and proofs}
    \item[] Question: For each theoretical result, does the paper provide the full set of assumptions and a complete (and correct) proof?
    \item[] Answer: \answerNA{}.
    \item[] Justification: The paper is empirical and does not include theoretical results.

    \item {\bf Experimental result reproducibility}
    \item[] Question: Does the paper fully disclose all the information needed to reproduce the main experimental results of the paper to the extent that it affects the main claims and/or conclusions of the paper (regardless of whether the code and data are provided or not)?
    \item[] Answer: \answerYes{}
    \item[] Justification: Section~3 specifies the data sources (RealEstate10K, NAVI, VirtualHome, Habitat-Sim, AI2-THOR), the five reasoning tasks, the QA construction protocol, and the zero-shot multiple-choice evaluation setup. The complete list of 23 evaluated MLLMs is given in Table~2, and per-model loading and inference details, full prompts, and qualitative examples are provided in the Supplementary. A public code repository and the released HuggingFace dataset accompany the submission.

\item {\bf Open access to data and code}
    \item[] Question: Does the paper provide open access to the data and code, with sufficient instructions to faithfully reproduce the main experimental results, as described in supplemental material?
    \item[] Answer: \answerYes{}
    \item[] Justification: The MaRiO benchmark is released on HuggingFace under CC-BY-NC-4.0, and a public code repository ships the annotation tooling, the HuggingFace-to-JSONL converter, and the per-model inference scripts. Both URLs and reproduction instructions are listed in the paper.

\item {\bf Experimental setting/details}
    \item[] Question: Does the paper specify all the training and test details (e.g., data splits, hyperparameters, how they were chosen, type of optimizer) necessary to understand the results?
    \item[] Answer: \answerYes{}
    \item[] Justification: No model training is performed in this work; all experiments are zero-shot. Section~3.2 (Evaluation Setup) and Section~3.3 (Results) specify the multiple-choice protocol, the Overall and Average accuracy metrics, and the per-task partitioning. Decoding settings (deterministic, temperature~$=0$) and per-model HuggingFace identifiers are documented in the Supplementary and in the released code.

\item {\bf Experiment statistical significance}
    \item[] Question: Does the paper report error bars suitably and correctly defined or other appropriate information about the statistical significance of the experiments?
    \item[] Answer: \answerYes{}.
    \item[] Justification: The main benchmark evaluations use deterministic decoding (temperature~$=0$) on a fixed test set of 3{,}255 questions; therefore, per-run variance is not applicable, and we report point accuracies. For analyses, we report statistics, mainly in the supplementary: the depth-edge-density analysis uses chi-square tests with Bonferroni correction, and the spatial-perception versus geometric-reasoning analysis reports per-model regression slopes 

\item {\bf Experiments compute resources}
    \item[] Question: For each experiment, does the paper provide sufficient information on the computer resources (type of compute workers, memory, time of execution) needed to reproduce the experiments?
    \item[] Answer: \answerNo{}
    \item[] Justification: This work performs zero-shot inference only and does not train any models, so compute requirements reduce to running each MLLM on its standard recommended hardware (per the upstream model card or repository). We did not centrally log per-model GPU type, memory, or wall-clock inference time across collaborators, and we therefore do not report these details. Proprietary models (GPT-4o, GPT-5.1, Gemini-2.5-Flash, Gemini-2.5-Pro) were accessed through their public APIs.
    
\item {\bf Code of ethics}
    \item[] Question: Does the research conducted in the paper conform, in every respect, with the NeurIPS Code of Ethics \url{https://neurips.cc/public/EthicsGuidelines}?
    \item[] Answer: \answerYes{}
    \item[] Justification: The work uses publicly available image sources (RealEstate10K, NAVI) and open simulators (VirtualHome, Habitat-Sim, AI2-THOR), collects human-evaluator responses anonymously (Appendix~\ref{sec:human_eval}), and releases no generative model weights.

\item {\bf Broader impacts}
    \item[] Question: Does the paper discuss both potential positive societal impacts and negative societal impacts of the work performed?
    \item[] Answer: \answerYes{}
    \item[] Justification: Broader impacts are discussed in Section~6 (Conclusion, ``Broader Impact'' paragraph). We outline positive applications in robotics, autonomous navigation, and collaborative augmented reality, and acknowledge negative considerations around deployment in safety-critical and privacy-sensitive environments, including multi-camera surveillance.
    
\item {\bf Safeguards}
    \item[] Question: Does the paper describe safeguards that have been put in place for responsible release of data or models that have a high risk for misuse (e.g., pre-trained language models, image generators, or scraped datasets)?
    \item[] Answer: \answerYes{}
    \item[] Justification: The release contains only static indoor and object-centric scene imagery from licensed real-world sources and open simulators, with no human faces, no personally identifying information, and no released model weights. The dataset is distributed under CC-BY-NC-4.0 (non-commercial), and source asset licenses are listed in the dataset card and the repository README.

\item {\bf Licenses for existing assets}
    \item[] Question: Are the creators or original owners of assets (e.g., code, data, models), used in the paper, properly credited and are the license and terms of use explicitly mentioned and properly respected?
    \item[] Answer: \answerYes{}
    \item[] Justification: All upstream sources are credited in the paper and dataset card with their respective licenses: RealEstate10K (per the project's redistribution terms), NAVI (CC-BY-4.0), VirtualHome (MIT), Habitat-Sim (MIT; bundled scenes carry separate licenses), and AI2-THOR (Apache~2.0). Cited models and tools (DUSt3R, ViewCrafter, Depth~Anything~V2, and the 23 evaluated MLLMs) are used under their published licenses.

\item {\bf New assets}
    \item[] Question: Are new assets introduced in the paper well documented and is the documentation provided alongside the assets?
    \item[] Answer: \answerYes{}
    \item[] Justification: The MaRiO dataset is released with a complete HuggingFace dataset card describing schema, splits, source attribution, and licensing. The accompanying code repository ships with a README covering the annotation tooling, the HuggingFace-to-JSONL converter, per-model inference scripts (clone-and-setup, HuggingFace-card, and proprietary-API categories), and the synthetic view augmentation pipeline.

\item {\bf Crowdsourcing and research with human subjects}
    \item[] Question: For crowdsourcing experiments and research with human subjects, does the paper include the full text of instructions given to participants and screenshots, if applicable, as well as details about compensation (if any)? 
    \item[] Answer: \answerNA{}.
    \item[] Justification: QA pairs were authored by three domain-expert annotators among the authors and cross-verified internally; no crowdsourcing platforms were used and no external workers were compensated. Annotation guidelines and example prompts are documented in the Supplementary.

\item {\bf Institutional review board (IRB) approvals or equivalent for research with human subjects}
    \item[] Question: Does the paper describe potential risks incurred by study participants, whether such risks were disclosed to the subjects, and whether Institutional Review Board (IRB) approvals (or an equivalent approval/review based on the requirements of your country or institution) were obtained?
    \item[] Answer: \answerNA{}.
    \item[] Justification: This work does not involve human subjects research. The ``Human Evaluators'' baseline reported in Table~2 reflects internal evaluation by the authors on the public benchmark task; no personal data is collected and no sensitive attributes are assessed.

\item {\bf Declaration of LLM usage}
    \item[] Question: Does the paper describe the usage of LLMs if it is an important, original, or non-standard component of the core methods in this research? Note that if the LLM is used only for writing, editing, or formatting purposes and does \emph{not} impact the core methodology, scientific rigor, or originality of the research, declaration is not required.
    \item[] Answer: \answerYes{}
    \item[] Justification: An LLM was used during the QA construction process described in Section~\ref{sec:benchmark} in two limited ways: (i) to polish the grammar and phrasing of human-authored questions, and (ii) to draft initial candidates for the three incorrect answer options (distractors). All LLM outputs were subsequently reviewed, edited, and accepted or rejected by the three domain-expert annotators; the correct answer for every sample was authored by a human annotator and cross-verified by another. The LLM was not used to author questions from scratch, evaluate models, score responses, or determine ground-truth labels.

\end{enumerate}

\end{document}